\documentclass{article}

\PassOptionsToPackage{numbers,compress}{natbib}
 \usepackage[preprint]{neurips_2026}

\usepackage[utf8]{inputenc} 
\usepackage[T1]{fontenc}    
\usepackage{hyperref}       
\usepackage{url}            
\usepackage{booktabs}       
\usepackage{amsfonts}       
\usepackage{nicefrac}       
\usepackage{microtype}      
\usepackage{xcolor}         
\usepackage{lipsum}
\usepackage{graphicx}
\usepackage{multirow}
\usepackage[table]{xcolor}
\definecolor{lightblue}{RGB}{193,236,250} 
\usepackage{wrapfig}
\usepackage{float}
\usepackage{tcolorbox}
\usepackage{listings}
\title{Covering Human Action Space for Computer Use: Data Synthesis and Benchmark}

\makeatletter
\def\thanks#1{\protected@xdef\@thanks{\@thanks
        \protect\footnotetext{#1}}}
\makeatother

\author{
    Miaosen Zhang$^{1}$\footnotemark[2] \thanks{$\dagger$ The work is completed during internship at Microsoft Research Asia.} \ \ 
    Xiaohan Zhao$^{2}$\footnotemark[2] \ \ 
    Zhihong Tan$^{3}$\footnotemark[2] \ \ 
    Huoshen Zhou$^{1}$ \ \ 
    Yijia Fan$^{4}$ \ \ 
    Yifan Yang$^{5}$ \AND 
    Kai Qiu$^{5}$ \ \
    Bei Liu$^{5}$ \ \ 
    Justin Wagle $^{5}$ \ \ 
    Chenzhong Yin$^{5}$ \ \ 
    Mingxi Chen$^{5}$ \ \ 
    Ji Li$^{5}$ \ \ 
    Qi Dai$^{5}$\footnotemark[3] \AND
    Chong Luo$^{5}$ \ \ 
    Xu Yang$^{1}$ \ \ 
    Xin Geng$^{1}$\footnotemark[3] \ \ 
    Baining Guo$^{1}$\footnotemark[3] \thanks{$\ddagger$ Corresponding authors.} 
    \vspace{0.4cm}\\
    {$^1$Southeast University} \quad 
    {$^2$Mohamed bin Zayed University of Artificial Intelligence} \\
    {$^3$Wuhan University} \quad 
    {$^4$Sun Yat-sen University} \quad
    {$^5$Microsoft}\\
    \texttt{\{miazhang,xgeng,307000167\}@seu.edu.cn} \quad 
    \texttt{qid@microsoft.com}
    }

\begin{document}

\maketitle

\begin{figure}[h]
    \centering
    \vspace{-1cm}
    \includegraphics[width=\textwidth,trim=150 150 150 100,clip]{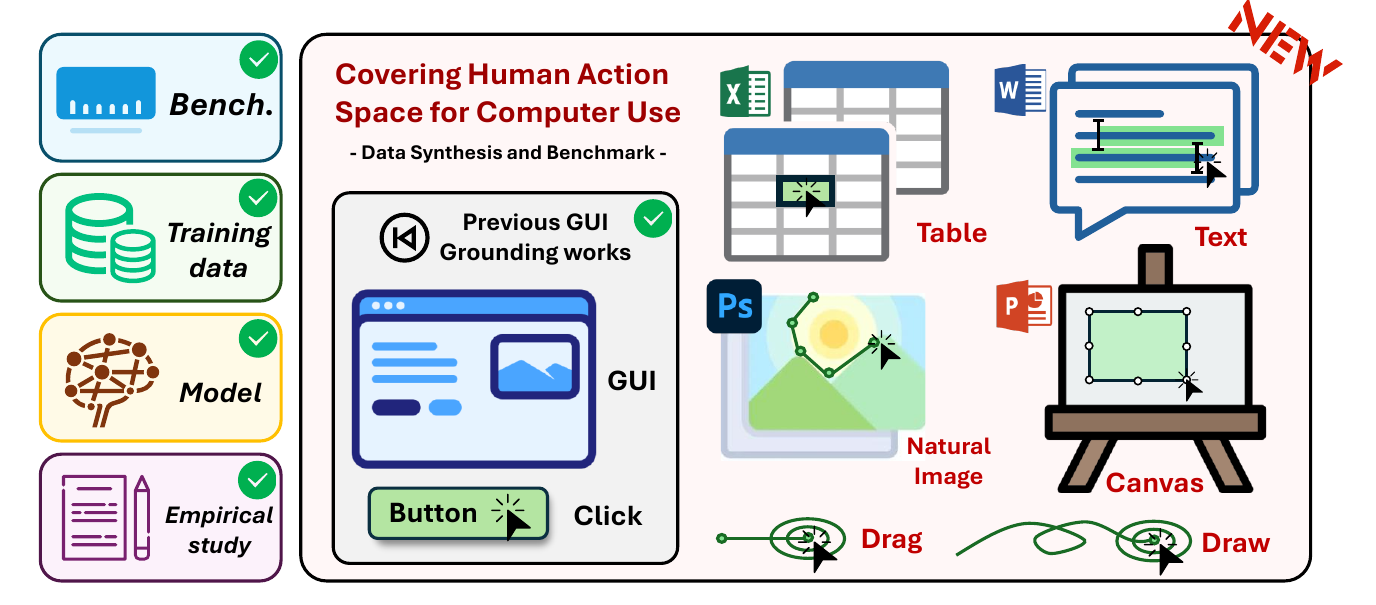}
    \caption{Overview. Prior GUI grounding research (lower-left panel of the inset) is dominated by click actions on standard GUI widgets. Computer-use agents, however, need to operate across a broader action space including editing tables, manipulating text, drawing on canvases, annotating
images, and execute richer actions including dragging and freehand drawing. We study this gap through four contributions: benchmark, data-synthesis pipeline, training models, and empirical studies.
}
    \label{fig:main-abs}
\end{figure}

\begin{abstract}
Computer-use agents (CUAs) automate on-screen work, as illustrated by GPT-5.4 and Claude. Yet their reliability on complex, low-frequency interactions is still poor, limiting user trust. Our analysis of failure cases from advanced models suggests a long-tail pattern in GUI operations, where a relatively small fraction of complex and diverse interactions accounts for a disproportionate share of task failures. We hypothesize that this issue largely stems from the scarcity of data for complex interactions. To address this problem, we propose a new benchmark \textbf{\textit{CUActSpot}} for evaluating models’ capabilities on complex interactions across five modalities: GUI, text, table, canvas, and natural image, as well as a variety of actions (click, drag, draw, etc.), covering a broader range of interaction types than prior click-centric benchmarks that focus mainly on GUI widgets. We also design a renderer-based data-synthesis pipeline: scenes are automatically generated for each modality, screenshots and element coordinates are recorded, and an LLM produces matching instructions and action traces. After training on this corpus, our  \textbf{\textit{Phi-Ground-Any-4B}} outperforms open-source models with fewer than 32B parameters. We will release our benchmark, data, code, and models at \hyperlink{https://github.com/microsoft/Phi-Ground.git}{https://github.com/microsoft/Phi-Ground.git}.
\end{abstract}

\section{Introduction}
Computer-Using Agent (CUA)~\citep{anthropic2024computeruse,openai2025cua} is a key direction for liberating human labor in digital work and enhancing productivity. CLI-based and GUI-based paradigms constitute two major interaction modes for CUAs. Compared with CLI-based CUAs, GUI-based CUAs inherently offer near-zero-cost cross-platform generalization, more user-friendly human–agent collaboration, and a higher theoretical ceiling: in principle, any computer task that humans can accomplish could also be completed by GUI-based CUAs.
However, owing to their efficiency and LLM-friendly interaction format, CLI-based CUAs~\citep{yang2024swe,wang2024openhands,agashe2025agent} have already demonstrated practical applicability faster than GUI-based.
Ideally, future CUAs will evolve into hybrid systems that combine the efficiency of CLI-based interaction with the flexibility and freedom of GUI-based operation. This paper primarily investigates the practical bottlenecks that hinder the deployment of GUI-based CUAs in real-world applications.

\begin{figure}[h]
    \centering
\includegraphics[width=\textwidth,trim=70 130 110 110,clip]{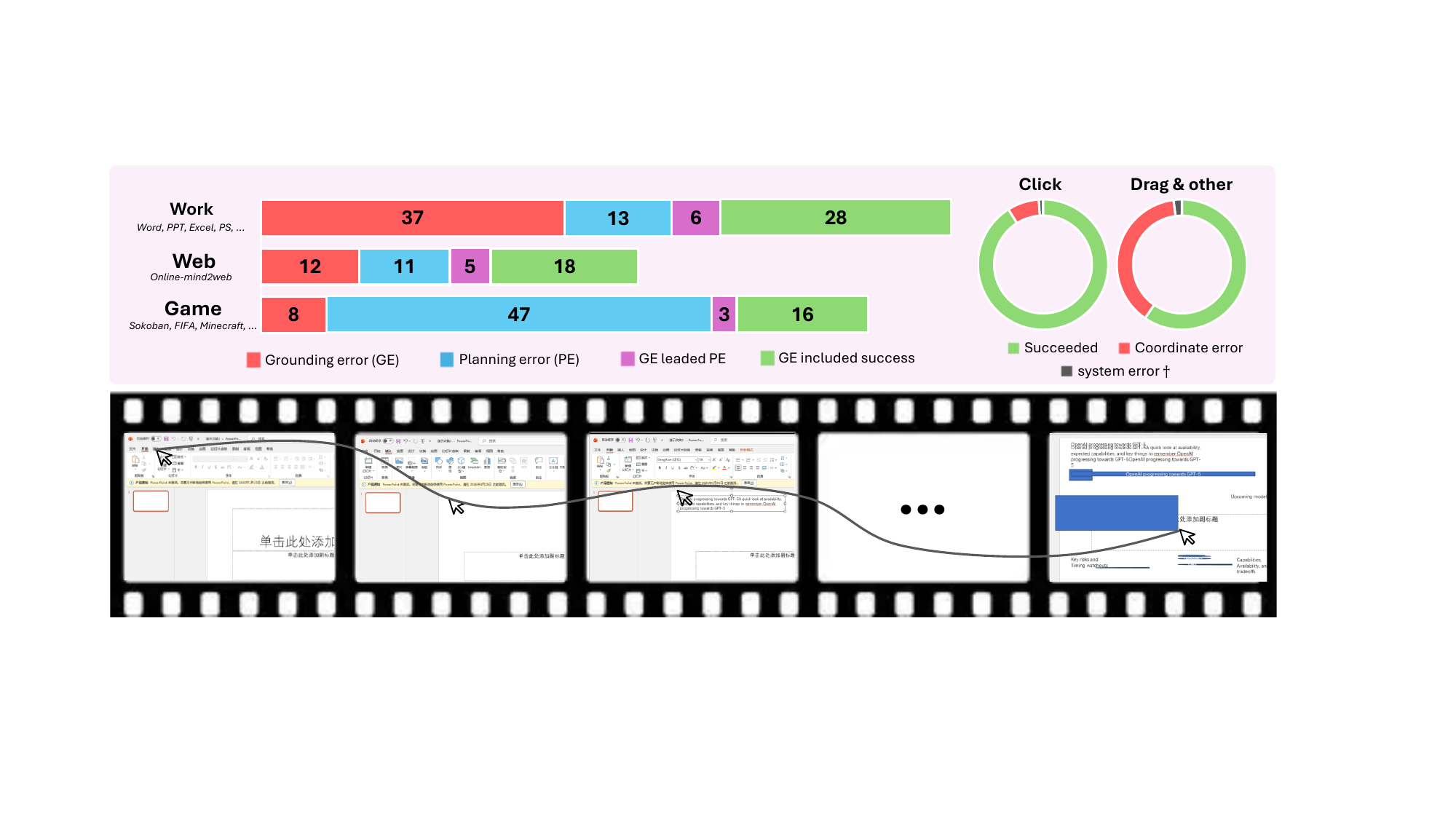}
\vspace{-0.4cm}
    \caption{Upper: Failure studies of GPT-5.4 computer use. $\dagger$: System errors refer to failures arising from the stochastic instability or limited robustness of certain VMs and interaction tools (e.g., PyAutoGUI). Lower: An E2E bad example of GPT-5.4 making a PPT introducing itself.}
    \label{fig:intro}
    \vspace{-0.2cm}
\end{figure}

We begin with a user study of GPT-5.4's~\cite{openai2026gpt54} computer-use capability on the Azure OpenAI platform. We collected nearly 200 tasks spanning three scenarios: work, web usage~\citep{xue2025illusion}, and gaming~\citep{zhang2026magebench, liu2024visualagentbench}, and executed them in a Windows VM, analyzing all failure cases that except system errors. As summarized in the upper part of Figure \ref{fig:intro}, 
we find that \textbf{\textit{Action Grounding}}~\cite{gou2024navigating,wu2024atlas,zhang2025phi,qin2025ui} is the most important source of error in the work setting, which is also the scenario users care about most.

In the past years, several challenging GUI grounding benchmarks~\cite{cheng2024seeclick,li2025screenspot,nayak2025ui,xie2024osworld} have emerged. However, the challenges these benchmarks emphasize do not align with those CUAs face in real-world settings. Existing benchmarks are often difficult because they involve rare high-resolution interfaces or require substantial software-specific knowledge~\cite{li2025screenspot,nayak2025ui}; yet their tasks are typically limited to single-click actions, and their targets are primarily GUI widgets.
In contrast, our empirical observations show that CUAs frequently need to operate on objects such as tables, documents, charts, and images, often through more complex interactions including dragging and drawing~\cite{nayak2025ui}. 
This mismatch has, in turn, influenced the direction of model development~\citep{gou2024navigating,wu2024atlas,zhang2025phi,liu2025infigui,gu2025ui,tang2025gui,liu2026infigui,zhou2025mai,xu2026mobile}: as shown in the Figure \ref{fig:intro}, the failure rate for complex interactions is far higher than that for simple clicking.

We therefore identify two major bottlenecks in the current development of GUI-based CUAs: the lack of benchmarks for evaluating complex operations and the lack of large-scale datasets for such interactions. To address these issues, we first manually construct \textbf{\textit{CUActSpot}}, a benchmark that covers a broad set of mouse-based actions that are common in computer-use workflows. It spans five modalities: GUI, Text, Table, Canvas, and Natural Image, and includes not only clicking, but also dragging and drawing actions, such as tracing object boundaries in Photoshop for image cutout.
We find that performance on CUActSpot differs substantially from conventional GUI grounding benchmarks~\cite{cheng2024seeclick,li2025screenspot,nayak2025ui,xie2025scaling}, while showing closer agreement with end-to-end agentic results such as OSWorld~\cite{xie2024osworld}. This suggests CUActSpot may better reflect real-world computer-use scenarios.

We further propose a data synthesis pipeline that obtains screenshots and coordinate-related metadata through code-based rendering, and we find that advanced GPT models can be leveraged to synthesize data for complex operations. Using this approach, we generate 50M samples that can support model pre-training or mid-training. We conduct ablation studies and empirical analyses over different data compositions and derive several insights. For example, we observe that, compared with simply scaling the amount of training data within a single modality, increasing data diversity substantially improves the model’s general interactive capability, a phenomenon we term \textbf{\textit{variety scaling}}. Finally, our trained and open-sourced \textbf{\textit{Phi-Ground-Any-4B}} achieves state-of-the-art performance among grounding models below 32B parameters. We hope that the benchmark, model, data, and insights presented in this paper will be valuable to the community and the broader industry.

\section{Related Works}

\paragraph{Computer Use Agents}
Computer-use agents (CUAs) perceive screens and perform actions (e.g., clicks and keystrokes) to complete tasks autonomously. CUA development follows two paradigms. \textit{Modular CUAs} pair a frontier VLM as a planner with a dedicated grounding model for precise low-level actions (e.g., UGround~\citep{gou2024navigating}, SeeClick~\citep{cheng2024seeclick}, OS-Atlas~\citep{wu2024atlas}), though the natural-language interface between them can lose spatial and contextual information. \textit{End-to-end CUAs} unify perception, reasoning, and action grounding within a single model, enabling joint optimization at the cost of massive training data. Commercial products such as Claude Computer Use~\citep{anthropic2024computeruse} and OpenAI CUA~\citep{openai2025cua} have brought this paradigm to end users, while open-source models including UI-TARS~\citep{qin2025ui}, OpenCUA~\citep{wang2025opencua} MAI-UI~\citep{zhou2025mai} and EvoCUA~\citep{xue2026evocua} have rapidly approached comparable performance. However, a substantial gap between CUAs and human performance persists in complex scenarios such as document editing or multi-application coordination~\citep{xie2024osworld}. A key contributor is action grounding.

\paragraph{GUI Action Grounding.}
GUI action grounding refers to localizing a target position on screen given a natural-language instruction, serving as a foundational capability for CUAs to execute precise actions. Early GUI agents decompose the screen into enumerable widgets (via accessibility trees, DOM, or Set-of-Marks) and prompt the model to select discrete IDs~\citep{zheng2024gpt,yang2023set,hong2024cogagent}. This paradigm naturally frames action grounding as a \textit{widget-centric, click-centric} task.

As data pipelines mature, the community has shifted to purely visual grounding, where models directly output screen coordinates ~\cite{wu2024atlas,cheng2024seeclick,qian2025uground,xu2024aguvis,lin2024showui,yang2025aria,wang2025ui}. Despite the shift, the widget-centric and click-centric prior persists: training data and evaluation benchmarks co-evolve along the same axis. On the data side, construction pipelines largely inherit the web-crawl and accessibility-tree paradigm, producing widget bounding boxes and click labels over tens of millions of elements. On the evaluation side, \textbf{\textit{grounding benchmarks}} share the same protocol: predict a single point from a natural-language instruction and check whether it falls within the target widget~\cite{wu2024atlas,cheng2024seeclick,li2025screenspot,nayak2025ui}. Notably, ScreenSpot-Pro~\cite{li2025screenspot} pushes difficulty toward high-resolution professional software with tiny targets, yet remains single-click on GUI widgets. Non-widget modalities such as tables, canvases, and natural images, and finer-grained operations like drawing, remain largely untouched. End-to-end agentic benchmarks~\citep{xie2024osworld,bonatti2024windows,rawles2024androidworld,zhou2023webarena,deng2023mind2web} involve richer interactions but measure task-level outcomes, making it difficult to isolate grounding as a factor. Across the field, the widget-and-click-centric prior remains pervasive. As a result, complex interactions beyond clicking remain undeserved in training and evaluation. As illustrated in Figure~\ref{fig:intro}, coordinate errors on such operations are far more frequent than on simple clicks, even for GPT-5.4~\citep{openai2026gpt54}.

\section{CUActSpot Benchmark}
In this section, we aim to evaluate models’ capabilities in handling complex GUI interactions. To this end, we introduce a new benchmark, CUActSpot. Compared with traditional GUI grounding tasks, CUActSpot features a broader range of more complex interaction types. At the same time, we reduce the amount of domain-specific knowledge required to complete the tasks, so that the evaluation results more accurately reflect a model’s action capabilities rather than overfitting to specialized knowledge. We begin by describing the metric used to compute the benchmark scores.
\subsection{Evaluation Rules and Metrics}
\begin{figure}[h]
    \centering
    \includegraphics[width=0.9\textwidth,trim=40 80 0 90,clip]{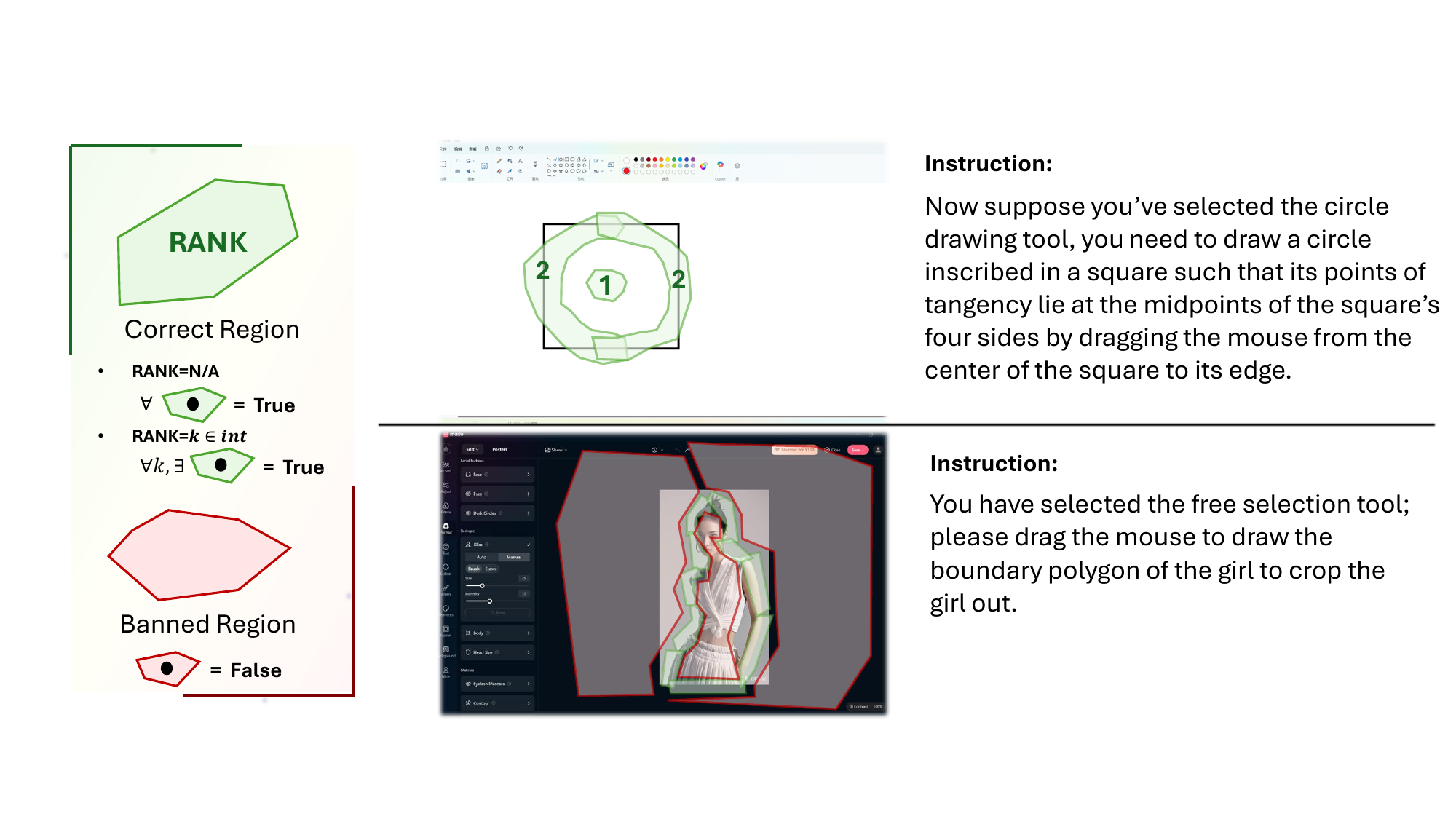}
    \caption{Benchmark evaluation rules and metric. More examples can be found in Appendix \ref{sec:app-bench-example}.}
    \label{fig:bench-rule}
    \vspace{-0.2cm}
\end{figure}

To evaluate various GUI interactions, including dragging, we first define two types of regions:
\begin{itemize}
    \item \textbf{Correct Region.} The coordinates predicted by the model (e.g., click locations or the start and end points of a drag) are required to lie within these regions, as shown by the green areas in Figure~\ref{fig:bench-rule}. A correct region may optionally have a rank attribute, which is used to evaluate order-sensitive actions. For instance, dragging along an arrow is order-sensitive, whereas dragging to select a span of text is order-insensitive, since the selection can be made by dragging either from front to back or from back to front.
    \item \textbf{Banned Region.} The model’s predicted actions must not occur within these regions. The purpose of introducing banned regions is to prevent metric gaming in tasks with N key points, where a model might otherwise click randomly across the entire screen in an attempt to inflate its score.
\end{itemize}
The dataset guarantees that, for each sample, the Correct Regions either all have a rank attribute or all lack one. In addition, some samples include Banned Regions, while others do not. Based on the above definitions of the two region types, we establish the following evaluation rules to determine whether a sample is considered correct, with priority applied in the order listed below.

\begin{itemize}
    \item \textbf{Rule 1.} If a sample defines any Banned Region, then the sample is marked as incorrect as soon as any coordinate predicted by the model (e.g., for a drag or a click) falls within a banned region.
    \item \textbf{Rule 2.} If the Correct Regions are ordered, then correctness is determined as follows: for each rank (where a given rank may correspond to one or more regions), it is sufficient for a key point to fall within any one of the regions associated with that rank; moreover, the sequence of predicted key points must match the order of the ranks. For example, in the upper-right example of Figure 3, dragging from the center outward to draw a circle is an order-sensitive action, but the model only needs to drag to any location on the circle’s radius for the action to be considered correct.
    \item \textbf{Rule 3.} If the Correct Regions are unordered, then the prediction is considered correct as long as each correct region contains at least one key point.
\end{itemize}

We determine whether each sample is successful according to the above rules, and report the sample success rate as the evaluation metric.

\subsection{Benchmark Statistics}

The entire construction pipeline of the CUActSpot benchmark was carried out manually. We first categorized GUI interaction targets into five common types:
\textbf{\textit{``GUI''}} refers to standard GUI widgets, such as buttons, checkboxes, and search bars.
\textbf{\textit{``Text''}} refers to operations performed directly on text, such as insertion and selection, which are common in applications like Microsoft Word and Notepad. Note that clicking a button containing text does not fall into this category.
\textbf{\textit{``Table''}} mainly refers to spreadsheet-style operations, as exemplified by Excel. In addition to clicking cells, actions such as dragging cell borders or corners are also included in this category.
\textbf{\textit{``Canvas''}} primarily refers to operations on graphical objects, as in PowerPoint.
\textbf{\textit{``Natural Image''}} refers to interactions within natural images, as in Photoshop, including clicking or dragging over specific image regions—for example, adjusting curves or drawing boundaries for image cutout.

\begin{table}[htbp]
  \centering
  \caption{Benchmark statistic comparison. The last row refer to the training dataset generated from our data synthesis pipeline in Section \ref{sec:data-pipe}. More details about how tasks and detailed tasks are classified can be found in Appendix \ref{sec:app-detailed-tasks}.}
  \resizebox{\textwidth}{!}{
    \begin{tabular}{lp{10em}p{8em}cc}
    \toprule
    \textbf{Works} & \multicolumn{1}{l}{\textbf{Modal}} & \multicolumn{1}{l}{\textbf{Action types}} & \textbf{\# Tasks} & \textbf{\# Detailed tasks} \\
    \midrule
    Prevs~\citep{wu2024atlas,cheng2024seeclick,li2025screenspot,nayak2025ui} & \multicolumn{1}{l}{1-3 (GUI, Table, Canvas)} & \multicolumn{1}{l}{click} & $\leq 3$     & $\leq 5$ \\
    \midrule
    \textbf{CUActSpot (ours)}  & \textbf{5} (GUI, Text, Table, Canvas, Natual Image) & click (1p), drag (2p), draw (Np) & \textbf{12}    & \textbf{33} \\
    \midrule
    \textbf{Data Syn. (ours)}  & \textbf{5} & click, drag, draw & 11 & 20 \\ 
    \bottomrule
    \end{tabular}
    }
  \label{tab:bench-statistic}%
\end{table}%

For each category, we further refined the task space according to the number of key points involved: one point (click), two points (drag), or 
N points (draw), as well as whether the action is ordered or unordered. Through iterative brainstorming, combined with realistic operations commonly performed in various software applications, we ultimately collected a diverse set of tasks, as summarized in Table \ref{tab:bench-statistic}.
After the tasks were collected and annotated, we further asked three additional individuals, independent of the original annotator, to attempt them. We then revised any ambiguous task descriptions and removed all tasks that could not be completed by humans. The final dataset contains 206 diverse and complex samples.

Comparing with existing GUI grounding benchmarks, our CUActSpot has the following uniqueness:
\begin{itemize}

    \item \textbf{Diverse task types. }Traditional benchmarks typically contain only click-based tasks, with targets largely limited to standard GUI elements, along with a small number of shapes or table cells. In contrast, our benchmark covers a much broader range of task types. Moreover, if we further distinguish tasks by the specific interaction target (see ``\# detailed tasks'' in Table \ref{tab:bench-statistic} for example, clicking an icon button and clicking a text button belong to the same high-level task type but correspond to different detailed tasks), the diversity of our benchmark becomes even greater.
    \item \textbf{Reduced ambiguity and reduced reliance on specialized knowledge. } In challenging benchmarks such as ScreenSpot-Pro, many samples are difficult even for humans to click correctly. This is partly because of the high screen resolution and occasional ambiguity in task descriptions, and partly because many samples require domain-specific software knowledge to determine the correct target. While such expertise is certainly relevant to CUA, it also introduces a potential confound: model performance may be influenced by how well the model is fitting to a particular software environment, rather than reflecting its grounding ability itself. We will further discuss this issue in the experimental section.
\end{itemize}

\section{General Action Grounding Data Synthetic Pipeline}
\label{sec:data-pipe}
\subsection{General Synthetic Pipeline}

\begin{figure}[h]
    \centering
    \includegraphics[width=\textwidth,trim=75 90 80 10,clip]{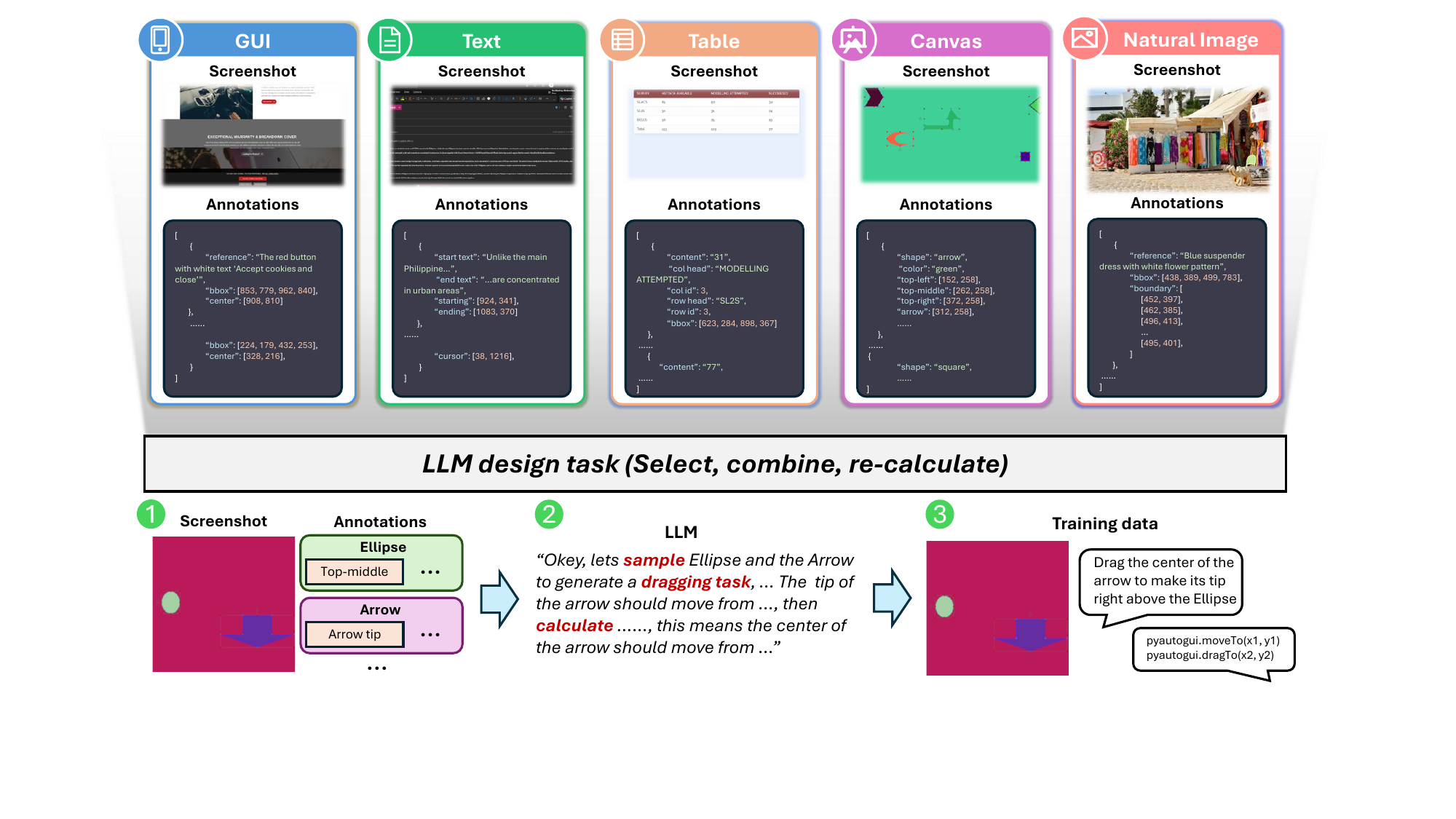}
    \vspace{-0.5cm}
    \caption{General data synthesis pipeline.}
    \label{fig:data-synpipe}
    \vspace{-0.3cm}
\end{figure}
To address the lack of training data for complex operations in CUA, we propose a fully synthetic data generation approach. Figure \ref{fig:data-synpipe} illustrates the overall synthesis framework. For each modality, we identify a code-based tool that can render screenshots. Because the visual elements (i.e., buttons in GUIs, cells in tables, and individual letters or characters in text) are generated through rendering, the same tool can also extract detailed coordinate information for each element, including bounding boxes and shape control points. Through a modality-specific pipeline, we obtain pairs consisting of a screenshot and a structured set of multiple elements together with their corresponding spatial metadata. We then design appropriate prompts to enable an LLM to select salient information from these element sets, combine them, and synthesize complex GUI operation tasks.

In the following subsections, we describe the rendering details and provide data examples for each modality. In practice, we design a separate system prompt for each modality (see Appendix \ref{sec:app-data-details}) and use the OpenAI o3~\citep{openai2025o3} model to generate tasks from the synthesized data. We not only allow the model to directly use the coordinate information provided in the annotations, but also permit it to perform intermediate calculations in order to construct more sophisticated tasks. We find that o3 performs this process effectively. For example, in the case shown at the bottom of Figure \ref{fig:data-synpipe}, Step 2 is an illustrative reconstruction written by us, since o3 does not disclose its chain-of-thought. Suppose a Canvas screenshot contains shapes such as an arrow and an ellipse. When all relevant element information is provided to the LLM, we observe that it can reason over these coordinates and generate the task shown in Step 3 after the necessary computation. Specifically, let the center of the arrow be $(x_1, y_1)$, the tip of the arrow be $(x_1, y_c)$, and the top control point of the ellipse be $(x_2, y_t)$. To make the arrow tip coincide with the top of the ellipse, the model infers that the arrow center should be moved from $(x_1, y_1)$ to $(x_2, y_2)$, where $y_2=y_t+y_1-y_c$. We observe many similar cases in practice, which substantially enriches the diversity of synthesized task types.

\subsection{GUI Element and Table Modal}
We use web-based tools to render both the GUI and table grounding datasets. For the GUI data, we reuse the data synthesis pipeline from Phi-Ground. In brief, we crawl webpages from CommonCrawl, then filter and clean them, render screenshots using the UI automation framework Playwright, and extract the bounding boxes of each button through JavaScript.

For the table data, we first collect tabular data in various formats, including LaTeX and Markdown, from Huggingface and convert them into HTML tables. We then employ an LLM to iteratively modify and evolve these tables, including changing their topology to introduce more complex structures such as multi-column layouts and merged cells, randomly masking a large number of cells, and revising the table contents, resulting 500k unique tables. In parallel, we prompt the LLM to create $\sim 5000$ CSS templates based on various open-source CSS libraries, where each template corresponds to a distinct table appearance style. By further randomizing properties such as colors and font weights, each template is expanded into multiple CSS instances. Finally, by combining these CSS instances with the HTML tables and rendering them as webpages, we obtain a large collection of table images with diverse visual styles.
\subsection{Text and Canvas Modal}

Both the text and canvas datasets are rendered using Python-based graphics and image-processing techniques. For the text data, we download 2,500 open-source English fonts and manually capture or collect approximately 200 text-background images at different resolutions, 
such as blank Microsoft Word documents and screenshots of Notepad windows. 
Using the PyQt5 library, we render textual content (from Wikipedia and GitHub) onto the blank regions of these backgrounds with randomly sampled fonts, colors, sizes, and weights, while recording the coordinates of every individual character.

For the canvas data, we directly use the plt library. We reproduce 15 common shape types typically found in Microsoft PowerPoint, including auxiliary visual elements such as dashed selection borders and white circular control points that appear around selected shapes. These shapes are then randomly placed onto blank canvases, with their type, color, size, canvas background color, width, and height all sampled at random. Different shapes may require different forms of positional annotation; for example, triangles are annotated by the coordinates of their vertices. All such geometric information is recorded in the annotations.

\subsection{Natural Image Modal}
For natural images, we use data from SAM~\citep{kirillov2023segment}. For each image, we first randomly sample five regions. Because these regions do not come with sufficiently detailed captions, we use GPT-4o~\citep{hurst2024gpt} to generate fine-grained descriptions for each selected region. SAM itself provides the bounding box and segmentation mask for every region. Based on these masks, we apply the Suzuki–Abe contour extraction algorithm~\citep{suzuki1985topological}, followed by contour sampling, to obtain polygonal boundary curves. These annotations are primarily used to support operations such as object cutout and zigzag-mask editing in Photoshop-like scenarios. All of this information is then packaged into the annotations.

\section{Experiments and Evaluations}
\subsection{Training Details}
For the datasets introduced in the previous section, we generated about 5M samples for each modality, except for the GUI modality, for which we generated 30M samples. Since our data are primarily intended for the pre-training or mid-training stages of VLMs, we require a base model that has not been exposed to GUI-related pre-training. To this end, we adopt Phi-3.5-VL~\citep{abdin2024phi3}, a 4B-parameter VLM, as the backbone. We put the detailed hyper-parameters and data proportion in Appendix \ref{sec:app-train-details}.

\subsection{Benchmarks Studies}

 In Table \ref{tab:main-result}, we present the performance of our model alongside several well-known open-source models on our benchmark, as well as on ScreenSpot-Pro and UI-Vision. Note that, there are many other well-known GUI grounding models, such as GTA1~\citep{yang2025gta1}. However, because many prior studies do not provide sufficient documentation or code of their benchmark evaluation coding details, we report only the models for which we were able to successfully reproduce the benchmark scores reported in their papers within a margin of ±5\%.

\paragraph{The pros and cons of knowledge barrier} The two most widely adopted benchmarks at present, ScreenSpot-Pro and UI-Vision, each cover a large collection of commonly used desktop applications, and many of their grounding tasks can only be completed with corresponding software-specific knowledge. For example, consider the grounding task: “Click the dodge tool icon button in Photoshop.” Even for a human user, this task would be difficult to complete if they did not know what the dodge tool icon looks like. This design offers an obvious advantage: it evaluates grounding ability while simultaneously testing software knowledge.

However, this design also introduces a notable drawback: benchmark performance becomes dominated by software-specific knowledge rather than grounding ability itself. In other words, if solving a test case requires knowledge of a particular application, then the model must have been trained on data from that application, which encourages model development to focus on covering the software included in the benchmark rather than on learning genuinely generalizable grounding capabilities. As shown in Table \ref{tab:main-result}, UI-Vision and ScreenSpot-Pro differ in software coverage, and UI-Vision also uses lower screen resolutions. Yet because ScreenSpot-Pro was introduced earlier and has achieved greater visibility, we observe that many recent models exhibit a substantial performance gap between the two benchmarks ($>20$ points). By contrast, for earlier models such as OS-Atlas and UGround, as well as for GPT-5.4, this gap is markedly smaller.

This gap should not be interpreted as direct evidence of overfitting; it may also reflect differences in benchmark design, software coverage, and the training-data mixtures used by different models. More directly, we further fine-tuned our pretrained Phi-Ground-Any model by incorporating the common-software data used in Phi-Ground, which was collected through Bing Search and may overlap with both benchmarks. The fine-tuning process included only click-based tasks. The results show substantial gains on both benchmarks, while performance on CUActSpot instead declined. This further demonstrates the sensitivity of existing benchmarks to the distribution of the training data.

\begin{table}[t]
  \centering
  \caption{GUI Grounding models and their results on ScreenSpot-pro (SS-pro), UI-Vision (UI-V) and CUActSpot. For UI-V and SS-Pro, the scores listed were taken from the literature when available, otherwise, they were obtained through our own evaluation. $\Delta$ refers to the gap between SS-pro and UI-V. $*$: Models did not report / released before SS-pro. $\dagger$: Models reported both SS-pro and UI-V.}
  \resizebox{\textwidth}{!}{
    \begin{tabular}{lcccccccccc}
    \toprule
    \multicolumn{1}{l}{\multirow{2}[4]{*}{\textbf{Model }}} & \multicolumn{1}{c}{\multirow{2}[4]{*}{\textbf{Date}}} & \multicolumn{1}{c}{\multirow{2}[4]{*}{\textbf{SS-pro}}} & \multicolumn{1}{c}{\multirow{2}[4]{*}{\textbf{UI-V}}} & \multicolumn{1}{c}{\multirow{2}[4]{*}{\textbf{$\Delta$}}}  & \multicolumn{6}{c}{\textbf{CUActSpot}} \\
\cmidrule(lr){6-11}      &    &       &        &    & \multicolumn{1}{c}{\textbf{GUI}} & \multicolumn{1}{c}{\textbf{Text}} & \multicolumn{1}{c}{\textbf{Table}} & \multicolumn{1}{c}{\textbf{Canvas}} & \multicolumn{1}{c}{\textbf{Image}} & \multicolumn{1}{c}{\textbf{Overall}} \\
    \midrule
    Phi-Ground-4B-16C$\dagger$~\citep{zhang2025phi}  &  2025-07     &   38.0    &   24.5   &   13.5   &  5.3  &  6.2  &  6.2  &  4.7  &  2.4    &  \cellcolor{lightblue} 5.0 \\
    Uground-V1-2B$*$~\citep{gou2024navigating}   &  2024-10  & 27.1    &   12.8  &   14.3    &  10.5  &  0.0  &  9.4  &  6.2  &  0.0   & \cellcolor{lightblue} 5.2 \\
        Uground-V1-7B$*$~\citep{gou2024navigating}  &  2024-10   &  31.1  &   12.9  &   18.2   &  18.4  &  0.0  &  3.1  &  9.4  &  2.4  &      \cellcolor{lightblue} 6.7 \\
    OS-Atlas-Base-7B$*$~\citep{wu2024atlas}  &  2024-10   &   18.9    &  9.0    &   9.9    &   15.8  &  0.0  &  12.5  &  10.9  &  0.0   &  \cellcolor{lightblue} 7.8 \\
     InfiGUI-R1-3B~\citep{liu2025infigui} &  2025-04     &   45.2    &   22.0   &   23.2   &   23.7  &  3.1  &  9.4  &  7.8  &  0.0    &  \cellcolor{lightblue} 8.8 \\
    UI-Venus-Ground-7B~\cite{gu2025ui}   &  2025-08    &   50.8    &   26.5   &  24.3   &    23.7  &  3.1  &  18.8  &  9.4  &  0.0   &  \cellcolor{lightblue} 11.0  \\
     GUI-G$^2$-7B~\citep{tang2025gui}   &  2025-07  &   47.5    &    26.4 &   21.1   &    23.7  &  6.2  &  15.6  &  7.8  &  4.8     &  \cellcolor{lightblue} 11.6 \\
    MAI-UI-2B$\dagger$~\cite{zhou2025mai}   &  2025-12    &   57.4    &  30.3  &   27.1   & 18.4  &  3.1  &  18.8  &  12.5  &  9.5   &  \cellcolor{lightblue} 12.5 \\
    GUI-Owl-1.5-8B-Think~\citep{xu2026mobile}  &  2026-02   &    57.6   &   33.2   &  24.4  &  23.7  &  9.4  &  18.8  &  10.9  &  7.1    &  \cellcolor{lightblue} 14.0 \\
     MAI-UI-8B$\dagger$~\cite{zhou2025mai}   &  2025-12   &    65.8   &      40.7 &   25.1   &   26.3  &  18.8  &  18.8  &  7.8  &  4.8    &  \cellcolor{lightblue} 15.3 \\
     GUI-Owl-1.5-8B-Instruct~\citep{xu2026mobile}  &  2026-02  &    71.1   &   37.4   &    33.7    &    23.7  &  15.6  &  18.8  &  9.4  &  9.5    &  \cellcolor{lightblue} 15.4 \\
     UI-Venus-Ground-72B~\cite{gu2025ui}   &  2025-08   &   61.9    &   36.8    &   25.1   &   28.9  &  18.8  &  18.8  &  10.9  &  9.5    &  \cellcolor{lightblue} 17.4 \\
     InfiGUI-G1-7B~\cite{liu2026infigui}  &  2025-08   &   51.9    &     26.1  &   25.8   &   44.7  &  18.8  &  37.5  &  9.4  &  4.8   &  \cellcolor{lightblue} 23.0 \\
    EvoCUA-8B~\citep{xue2026evocua} &  2026-01    &   45.4    &   15.6    &   29.8  &  18.4  &  40.6  &  34.4  &  9.4  &  16.7  &  \cellcolor{lightblue}23.9  \\
    UI-TARS-1.5-7B~\citep{qin2025ui} &  2025-04 &    42.6   &  22.3  &   20.3   &   42.1  &  28.1  &  34.4  &  14.1  &  23.8   &  \cellcolor{lightblue} 28.5 \\
    EvoCUA-32B~\citep{xue2026evocua}  &  2026-01    &    49.8   &   20.9  &   28.9    &    28.9  &  31.2  &  40.6  &  25.0  &  16.7   &  \cellcolor{lightblue} 28.5 \\
    OpenCUA-7B~\cite{wang2025opencua}   &  2025-08  &  50.0   &    25.5  &    24.5   &  42.1  &  37.5  &  53.1  &  28.1  &  38.1   &     \cellcolor{lightblue} 39.8 \\
        \midrule
      \textbf{Phi-Ground-Any-4B (ours) $\dagger$}    &   2026-05  &  26.3  &  15.8   &   10.5   &   44.7    &   34.4    &  68.8     &  40.6  &   33.3    &  \cellcolor{lightblue}  44.4 \\
      \textbf{$+$ APP data finetuned $\dagger$}    & 2026-05  &  41.5  &  29.7  & 11.8   &   52.6    &   18.8    &  59.4     &  32.8  &   19.0    &  \cellcolor{lightblue}  36.5 \\
    \midrule
     OpenCUA-32B~\cite{wang2025opencua}   &  2025-08  &    55.3   & 26.3   &   29.0   &  55.3  &  46.9  &  68.8  &  39.1  &  52.4   &  \cellcolor{lightblue} 52.5 \\
    GPT-5.4 (Azure)$*$~\cite{openai2026gpt54}  &  2026-03    &   44.5    &   37.9   &    6.6    &   73.7  &  43.8  &  87.5  &  65.6  &  47.6   &  \cellcolor{lightblue} 63.6  \\
    
    \bottomrule
    \end{tabular}}
    \vspace{-0.6cm}
  \label{tab:main-result}%
\end{table}%

\begin{wraptable}{r}{0.4\textwidth}
    \vspace{-0.7cm}
    \centering
    \caption{OSWorld results \\ (max actions $=30$)}
    \resizebox{0.4\textwidth}{!}{
    \begin{tabular}{lp{6.275em}cc}
    \toprule
    \textbf{Planner} & \textbf{Grounder} & \textbf{SS-pro} & \textbf{OSWorld} \\
    \midrule
    GPT-5.4 & GUI-Owl-1.5-8B-Instruct~\citep{xu2026mobile} & 71.1  & 37.7 \\\midrule
    GPT-5.4 & MAI-UI-8B~\citep{zhou2025mai} & 65.8  & 38.2 \\\midrule
    GPT-5.4 & GPT-5.4~\citep{openai2026gpt54} & 44.5  & 44.1 \\\midrule
    GPT-5.4 & Phi-Ground-Any-4B &   26.3    &  42.4   \\
    \bottomrule
    \end{tabular}}
  \label{tab:osworld}
  \vspace{-0.3cm}
\end{wraptable}

\paragraph{Comparison with agentic benchmarks} Interestingly, we find that the top models on CUActSpot (i.e., GPT-5.4, OpenCUA, EvoCUA, UI-TARS) also happen to report results on OSWorld and treat OSWorld as major criterion. In our view, this does not suggest that CUActSpot and OSWorld can serve as substitutes for one another; rather, it reflects a form of statistical bias. Specifically, research efforts that genuinely focus on agentic settings and explicitly aim to optimize for them are also more likely to collect training data with broader modality coverage and more diverse interaction types. 

We further use OSWorld to evaluate grounding ability in Table \ref{tab:osworld}. During OSWorld evaluation, we uniformly employed GPT-5.4 to generate single-step natural-language instructions and required each grounder to predict the corresponding action parameters. In this way, planning was controlled across all methods, and grounding was the only variable. We selected two models whose performance on ScreenSpot-Pro was substantially higher than that of GPT-5.4 and Phi-Ground-Any-4B; however, their results on OSWorld did not show a correspondingly significant advantage. This finding suggests the notable mismatch with real-world scenarios.

\paragraph{The effects of data synthesis pipeline} The Phi-Ground-Any-4B model, trained on fully synthetic data together with OpenCUA data, demonstrates strong performance on CUActSpot, outperforming all open-source models smaller than 32B parameters. Although its performance on ScreenSpot-Pro and UI-Vision is relatively weak, our fine-tuning experiments on application-specific data from Phi-Ground indicate that this is primarily a consequence of differences in data distribution. After fine-tuning, the model surpasses Phi-Ground on both ScreenSpot-Pro and UI-Vision, thereby validating the effectiveness of the new data synthesis pipeline as a pretraining strategy.

\subsection{Empirical Studies and Ablations}
\label{sec:ablation}
\begin{figure}[t]
    \centering
    \includegraphics[width=\textwidth,trim=0 0 0 0,clip]{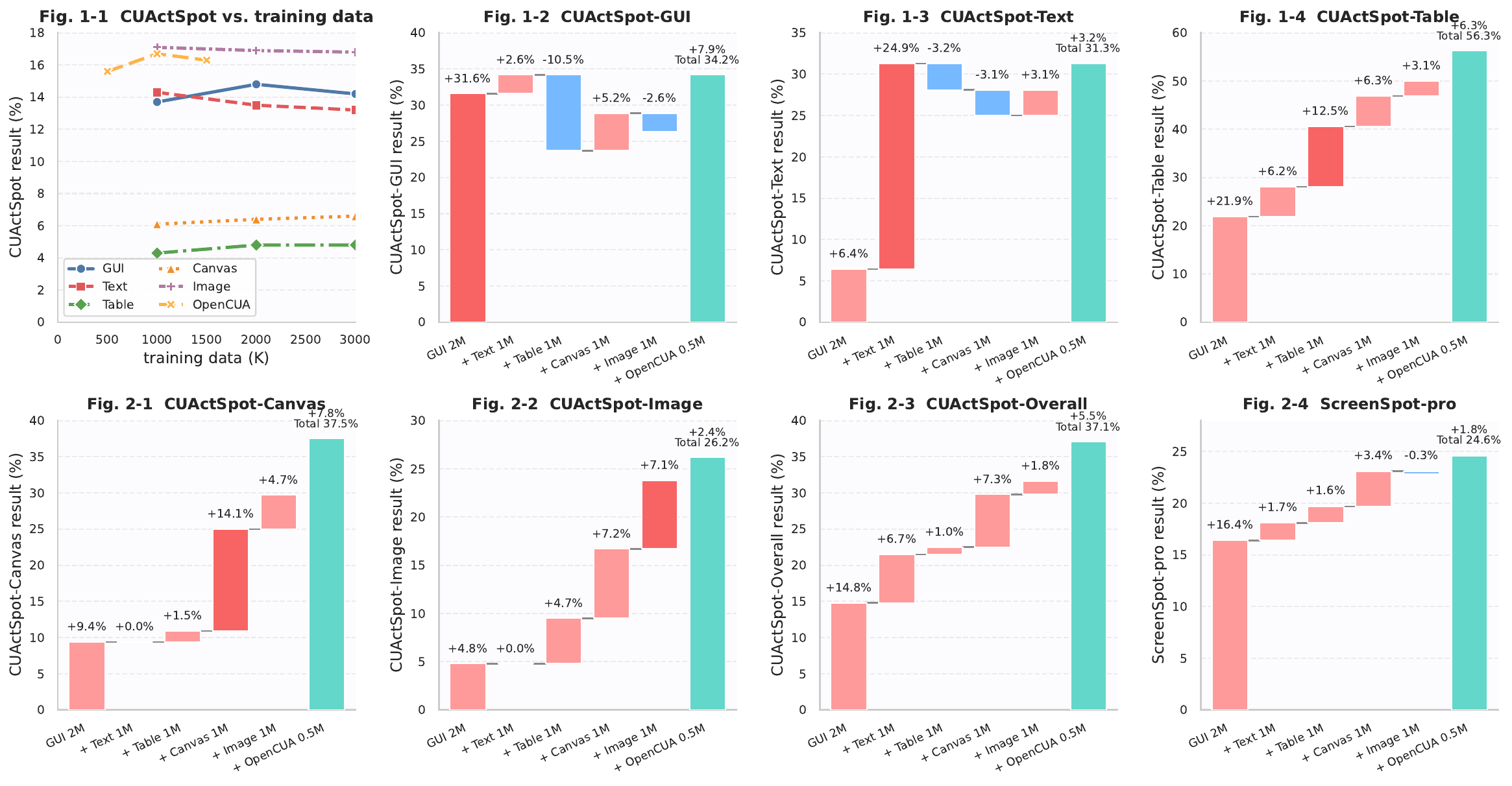}
    \vspace{-0.5cm}
    \caption{Data ablation results. Fig. 1-1: Independently scaling the training budget for each component in the training set shows that increasing the amount of training does not yield sustained improvements. Fig. 1-2 to 2-4: Waterfall plots illustrating the incremental effects of progressively adding different training datasets on each benchmark.}
    \label{fig:data-abl}
    \vspace{-0.5cm}
\end{figure}

We perform component-wise experiments and ablation studies on the data composition. As illustrated in Fig. \ref{fig:data-abl}, and with the detailed numerical results provided in Appendix \ref{sec:app-train-details}.

\paragraph{Variety scaling is the key} Our ablations suggest that scaling a single modality alone is less effective than increasing task and modality diversity in this setting. This is reminiscent of the fact that, in conventional vision tasks, we do not observe the kind of emergent intelligence seen in LLM. However, when we increase both the number of task types and the diversity of modalities, not only does performance on the corresponding modality improve (dark red bars), but capabilities on other modalities also improve gradually. Comparing Subfigures 1-1 and 2-3 suggests that the key variable driving continual learning is the diversity of task types. We therefore hypothesize that, for computer-use grounding, task diversity may be at least as important as raw data scale. To perform well across a wide range of tasks, the model must learn knowledge that generalizes across modalities.

\begin{wraptable}{r}{0.3\textwidth}
    \centering
    \vspace{-0.6cm}
    \caption{Task solved}
    \vspace{0.2cm}
\resizebox{0.3\textwidth}{!}{
    \begin{tabular}{lc}
    \toprule
     & \textbf{N Detailed Tasks} \\
    \midrule
    CUActSpot & 33 \\\midrule
    Data Syn. & 20 \\\midrule
    Model & 27 \\
    \bottomrule
    \end{tabular}}
  \label{tab:tasks-solved}
  \vspace{-0.3cm}
\end{wraptable}

\vspace{-0.2cm}
\paragraph{Cross-task generalization} We quantify the number of detailed tasks in CUActSpot that the trained model has the potential to accomplish, where a task is counted as feasible if the model successfully completes at least one sample from that task. 
Interestingly, the model succeeds on a larger number of detailed tasks than were explicitly present in the training set, suggesting limited compositional generalization across detailed tasks.
For example, a model that learns to interact with textual elements and to manipulate visual regions separately may subsequently acquire the ability to operate on text embedded within visual content, such as editing text inside a presentation figure or selecting text from natural images, even when such compositions are not explicitly present in the training data. We believe that, as the diversity of task types continues to grow, it will become increasingly feasible to train more general-purpose CUA systems.

\vspace{-0.2cm}
\section{Conclusions and Limitations}

In this work, we study the long-tail challenge in computer-use grounding and introduce CUActSpot, a benchmark that broadens evaluation beyond click-centric settings to more diverse interactions and modalities. We further present a scalable synthesis pipeline and show through ablations that increasing task and modality diversity is more effective than scaling a single modality in our setting. Empirically, our Phi-Ground-Any-4B achieves strong performance on complex interaction benchmarks and remains competitive among models of similar scale. Nevertheless, CUActSpot is a diagnostic benchmark with manually curated samples and does not exhaustively cover real-world workflows, especially long-horizon and stateful scenarios. In addition, while synthetic data enables broad and controllable coverage, improving alignment with real-world distributions remains an important direction for future work.


\newpage
\appendix

\section{CUActSpot Details}
\subsection{Detailed Tasks Breakdown}
\label{sec:app-detailed-tasks}

The following two tables present the specific task categories included in the CUActSpot benchmark. In constructing this benchmark, we first systematically decomposed the full range of mouse interactions that may arise during human computer use. For each fine-grained task, we collected corresponding operational data across several relevant software applications. As a result, most of the data consist of independent tasks. From the perspective of model training, clicking on a populated cell and clicking on a blank cell should be regarded as entirely different tasks: training a model on one of these tasks does not enable it to generalize to the other.

\paragraph{Error bars} According to our experiments, CUActSpot exhibits fluctuations of approximately ±3\% between adjacent checkpoints during training, while the variation observed during testing with different temperature settings is around 2\%. These results are comparable to those of ScreenSpot-pro, for which we also observed fluctuations of about 3\% during training and 1\% under different temperature settings.

\begin{table}[htbp]
  \centering
  \caption{Detailed tasks. Block number of `N key points' is the `\# Tasks' in Table \ref{tab:bench-statistic}. Row number is the `\# Detailed Tasks' in Table \ref{tab:bench-statistic}}
  \resizebox{\textwidth}{!}{
    \begin{tabular}{cccp{25em}}
    \toprule
    Modal & N key points & Target type & Example \\
    \midrule
    \multirow{6}[12]{*}{GUI} & \multirow{2}[4]{*}{1} & icon  & Click on the heart button of the sun glasses. \\
\cmidrule{3-4}          &       & text  & Click on the Warrios' score. \\
\cmidrule{2-4}          & \multirow{4}[8]{*}{2} & icon  & drag the Installer.exe into the file upload area on the Huggingface repository upload page.  \\
\cmidrule{3-4}          &       & slide bar & Drag the video progress bar to approximately 10 minutes and 15 seconds. \\
\cmidrule{3-4}          &       & empty region & Drag the top edge of the OneNote window to the right until its right side is flush with the right side of the screen. \\
\cmidrule{3-4}          &       & text  & Drag the Java17 folder into "java21" shown on the file path of the file manager. \\
    \midrule
    \multirow{5}[10]{*}{Text} & \multirow{2}[4]{*}{1} & between text & Click once at the position before "and the people" to set the cursor. \\
\cmidrule{3-4}          &       & empty region & I now need to add another method to the class; please click below the add function to position the cursor. \\
\cmidrule{2-4}          & \multirow{3}[6]{*}{2} & select text span & Drag the mouse to highlight all the text numbered 1-17. (including the numbers). \\
\cmidrule{3-4}          &       & One word & Drag the mouse to select the English word “Which” in the question in the cell in row 5 of the “Wednesday” column; this word is above the bolded “unkind.” \\
\cmidrule{3-4}          &       & drag text span & Drag the selected code snippet to the correct position. \\
    \midrule
    \multirow{6}[12]{*}{Table} & \multirow{2}[4]{*}{1} & empty cell & Please select cell H11 in the table. \\
\cmidrule{3-4}          &       & content cell & Please select the cell that contains "banana" in the table. \\
\cmidrule{2-4}          & \multirow{4}[8]{*}{2} & drag cell & Drag the selected cells to the next row. \\
\cmidrule{3-4}          &       & select cells & Drag the mouse to select the two cells containing apple and banana. \\
\cmidrule{3-4}          &       & edge  & Drag the right boundary of the selected table column header to make the column width four times its original size. \\
\cmidrule{3-4}          &       & corner & Drag the lower-right corner of cell I7 down to the lower-right corner of cell I13 to apply the formula and calculate the sums for the other six rows. \\
    \bottomrule
    \end{tabular}
}
  \label{tab:addlabel}%
\end{table}%

\begin{table}[t]
  \centering
  \caption{Detailed tasks. Block number of `N key points' is the `\# Tasks' in Table \ref{tab:bench-statistic}. Row number is the `\# Detailed Tasks' in Table \ref{tab:bench-statistic}}
    \resizebox{\textwidth}{!}{
    \begin{tabular}{cccp{25em}}
    \toprule
    Modal & N key points & Target type & Example \\
    \midrule
    \multirow{8}[16]{*}{Canvas} & 1     & shape & The circle layer I drew in WPS PPT is located beneath the fan-shaped layer. Please avoid selecting the fan or any other shapes and select only the circle layer. \\
\cmidrule{2-4}          & \multirow{5}[10]{*}{2} & empty region & Please drag the mouse to select the diamond and the heart shape inside it, but do not select any other shapes. \\
\cmidrule{3-4}          &       & point & Drag the control point on the red curve down slightly, but do not go below the diagonal. \\
\cmidrule{3-4}          &       & text  & I have selected a text box in WPS PPT; please drag it to the exact center of the slide. \\
\cmidrule{3-4}          &       & shape & In the diagrams document, drag node 2 to the exact midpoint between node 1 and node 2. \\
\cmidrule{3-4}          &       & line/arrow & I have selected node 1 in the diagrams document. Now please drag the blue arrow on its right and connect it to the left side of node 2. \\
\cmidrule{2-4}          & \multirow{2}[4]{*}{N} & point & Connect the black dots in the figure from smallest to largest. \\
\cmidrule{3-4}          &       & empty region & Assuming the polygon drawing tool is already selected. Detect the center points of all specified squares in the image and use these centers as polygon vertices.  \\
    \midrule
    \multirow{8}[16]{*}{Image} & \multirow{2}[4]{*}{1} & object & You are using the AI cutout feature and need to deselect the person on the far right. Just click once inside that person in the red-highlighted area on the left. \\
\cmidrule{3-4}          &       & region & Please click once on the grass outside the puppy's outline in the image. \\
\cmidrule{2-4}          & \multirow{4}[8]{*}{2} & image & In Photoshop, drag the image to the far right edge, keeping it at the same height. \\
\cmidrule{3-4}          &       & object & I created a small sun graphic; please drag it directly above the boat. \\
\cmidrule{3-4}          &       & point & I am using the image cropping feature to crop out the white boat in the frame without leaving any extra space. \\
\cmidrule{3-4}          &       & region & I have now selected the manual face- and waist-slimming feature. The woman’s face is not symmetrical, so I want you to drag from the inside of her left jawline outward to slightly enlarge the left side of her face. \\
\cmidrule{2-4}          & \multirow{2}[4]{*}{N} & zig-zag mask & I have now selected the eraser tool. Please drag it over the entire deer to select it. \\
\cmidrule{3-4}          &       & boundary & I have activated the free selection tool. Please drag the mouse to draw a boundary polygon around the tree’s reflection in the water at the bottom right of the image to select the reflected tree area. \\
    \bottomrule
    \end{tabular}}
  \label{tab:addlabel}%
\end{table}%

\subsection{Benchmark examples}
\label{sec:app-bench-example}
\begin{figure}[H]
    \centering
    \includegraphics[width=\textwidth,trim=0 0 0 0,clip]{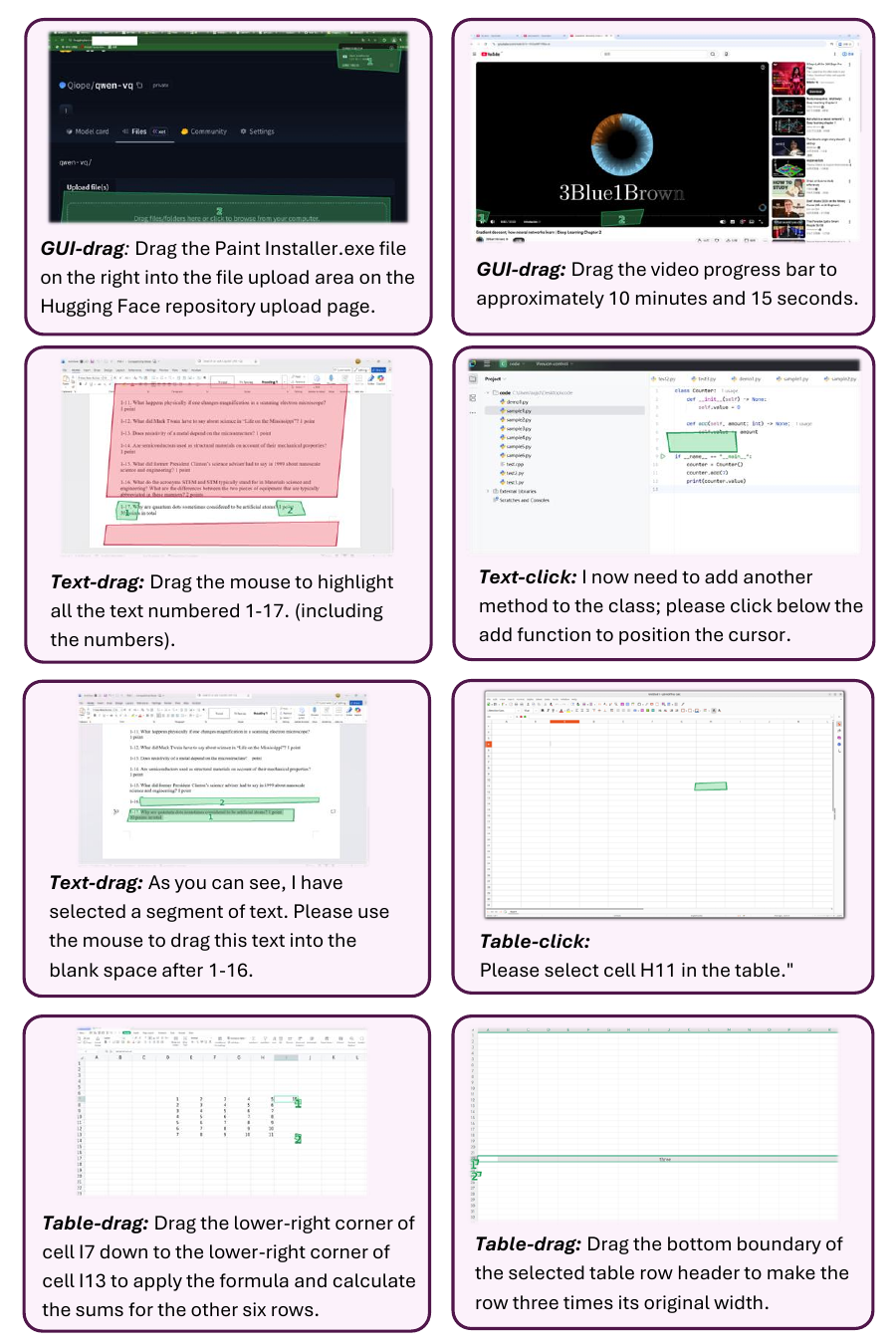}
    \caption{Examples of CUActSpot.}
    \label{fig:cuactspot-example1}
\end{figure}

\begin{figure}[H]
    \centering
    \includegraphics[width=\textwidth,trim=0 0 0 0,clip]{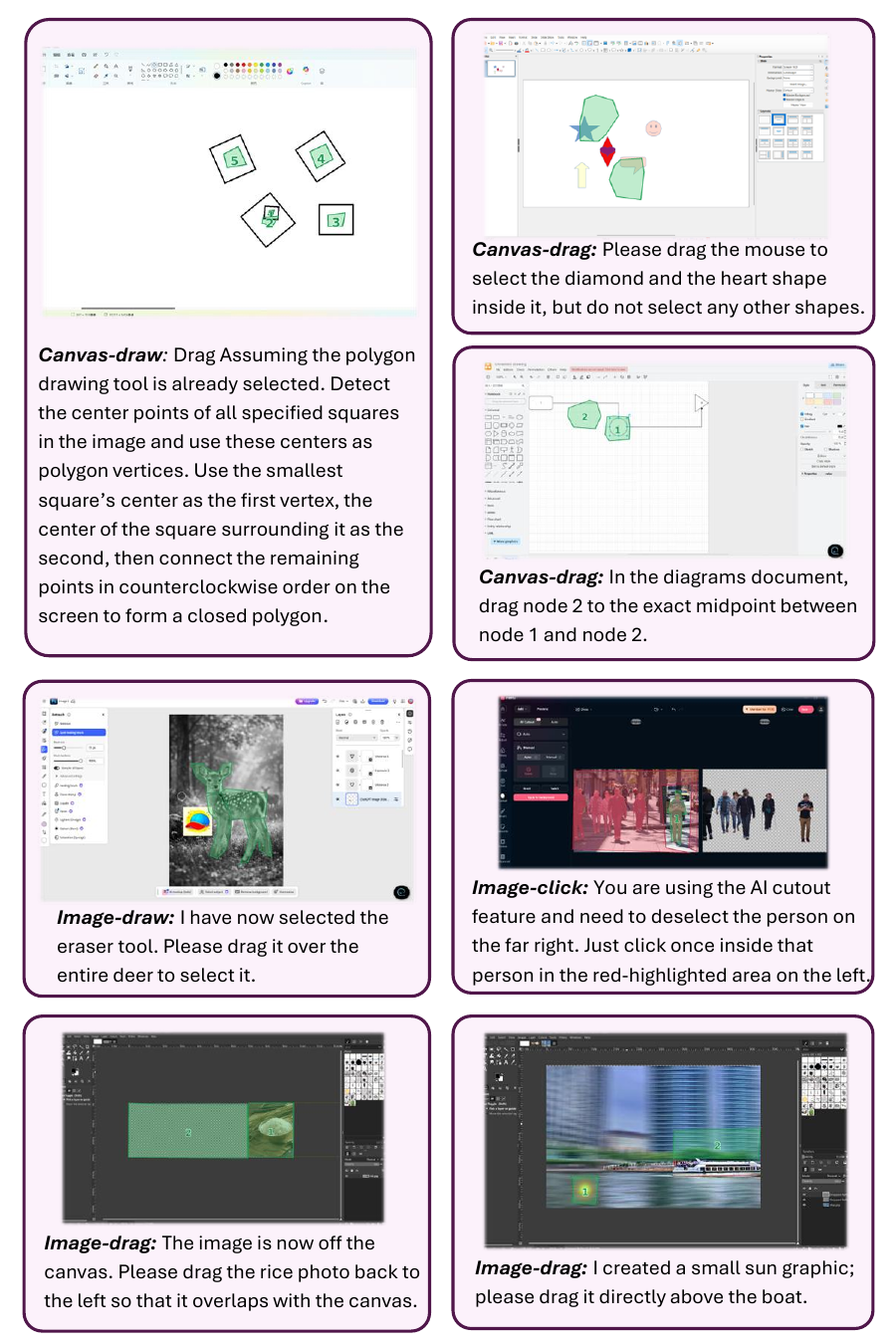}
    \caption{Examples of CUActSpot.}
    \label{fig:cuactspot-example1}
\end{figure}

\newpage

\section{More Training Details}
\label{sec:app-train-details}
\subsection{Data sampling}

During training, we fix the visual input to 16 crops and incorporate the various data augmentation strategies introduced in Phi-Ground~\citep{zhang2025phi}. For all experiments in this section, we use a fixed batch size of 5120 and a learning rate of $8\times 10^{-5}$. In addition, a weight decay of 0.01 and gradient clipping at 0.1 are important for maintaining training stability. 

In actual training and related experimental settings, we adopt the data composition shown in the following Table \ref{tab:data-samp}. We increased the proportion of OpenCUA data because it is manually annotated and therefore expected to be of higher quality. However, as shown in Section \ref{sec:ablation}, using only OpenCUA yields unsatisfactory performance due to its limited scale. The overall training budget is approximately 100B tokens. Training will take about 30 hours on 80 NVIDIA H100 GPUs.
\begin{table}[htbp]
  \centering
  \caption{Data proportion of Phi-Ground-Any model's training}
    \begin{tabular}{lcccc}
    \toprule
    \textbf{dataset type} & \textbf{samples} & \textbf{used samples} & \textbf{epoches} & \textbf{weight} \\
    \midrule
    GUI   & 30,432,242 & 6,800,000 & 0.22345 & 0.34 \\
    Text  & 6,083,400 & 5,000,000 & 0.82191 & 0.25 \\
    Table & 5,242,630 & 2,000,000 & 0.38149 & 0.1 \\
    Canvas & 4,323,253 & 2,000,000 & 0.46261 & 0.1 \\
    Image & 4,743,675 & 3,000,000 & 0.63242 & 0.15 \\
    OpenCUA & 340,665 & 1,200,000 & 3.52252 & 0.06 \\
    \bottomrule
    \end{tabular}%
  \label{tab:data-samp}%
\end{table}%

\subsection{Data ablation results}

All experiment results in this paper use a best-checkpoint strategy: we save checkpoints for every 100 training steps, and report the best checkpoint among them.

\begin{table}[htbp]
  \centering
  \caption{Data ablation results}
    \begin{tabular}{lcccccccc}
    \toprule
    \textbf{Data} & N Samples (M) & GUI   & Text  & Table & Canvas & Image & Overall & SSP\\
    \midrule
    GUI 2M   & 2     &   31.6   &   6.3    &   21.9    &   9.4    &   4.8    & 14.8  & 16.4  \\
    $+$ Text 1M  & 3     &  34.2    &   31.3    & 28.1     &   9.4    &   4.8    &  21.5 &  18.1  \\
    $+$ Table 1M  & 4     &   23.7    &   28.1   &   40.6  &   10.9    &   9.5    & 22.5 & 19.7  \\
    $+$ Canvas 1M & 5     &  28.9    &   25.0    &  46.9     &   25.0    &    16.7   &  28.5  &  23.1 \\
    $+$ Image 1M & 6     &   26.3   &   28.1   &  50.0     &   29.7    &   23.8    &  31.6  &  22.8   \\
    $+$ OpenCUA 0.5M & 6.5     &    34.2   &   31.3    & 56.3    &   37.5    &    26.2   & 37.1  &  24.6 \\
    \bottomrule
    \end{tabular}%
  \label{tab:data-abl-details}%
\end{table}%

\newpage
\section{Data Synthesis Details}
\label{sec:app-data-details}
\subsection{GUI}

\begin{figure}
    \centering
    \includegraphics[width=\linewidth, trim=40 90 30 10, clip]{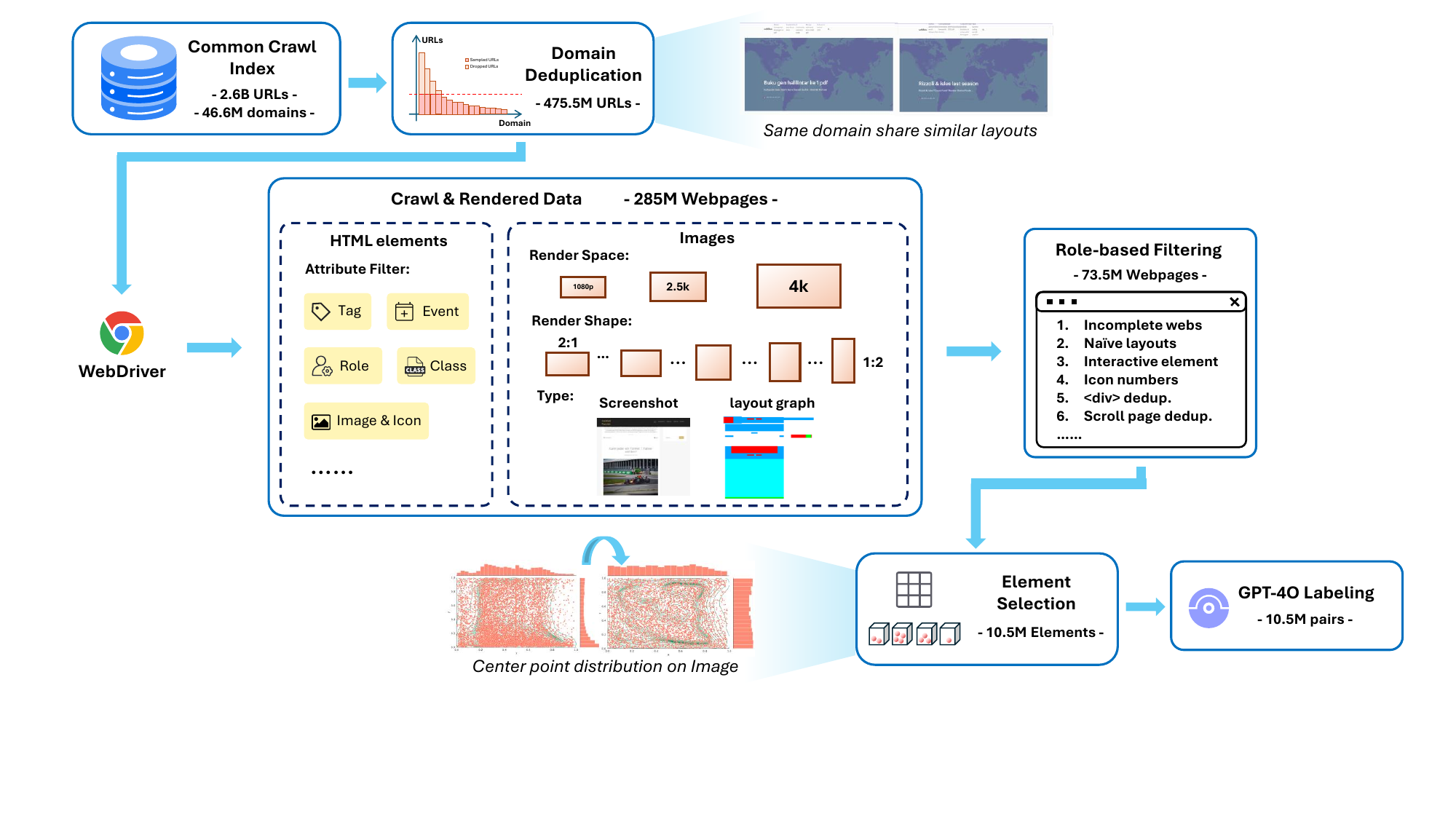}
    \caption{CommonCrawl data processing pipeline.}
    \label{fig:CC-cleaning}
\end{figure}

To acquire larger-scale data for better scaling up of training, we also obtained web pages from CommonCrawl and rendered screenshots to generate training data. However, the web data contained a significant amount of noisy data that caused training failures. To address this, we constructed a highly specific data cleaning pipeline, as illustrated in Figure \ref{fig:CC-cleaning}. Below are the detailed steps of each stage:

\paragraph{Index and domain deduplication} We utilized the \textit{CC-MAIN-2024-46} crawl from CommonCrawl. After a basic deduplication of URLs (exact match), filtering by language (retaining only English), and webpage status (retaining only 2xx, 301, and 302), we were left with 2.6 billion URLs. These 2.6B URLs originate from 45.6 million unique domains, with the number of pages from the same domain displaying a long-tail distribution. For instance, the largest domain contains 204K different pages. We observed that pages from the same domain exhibit strong consistency in layout. Therefore, to ensure the generalizability of our model, we performed random sampling so that no more than 50 pages were selected from each domain. After this round of sampling, we were left with 475.45M URLs.

\paragraph{Rendering} We utilized the Selenium library and Google Chrome Driver to render webpage screenshots. During the rendering process, we randomly selected from three different pixel areas corresponding to 1080p, 2K, and 4K screen resolutions. The aspect ratio of the images was randomly chosen between 2:1 and 1:2. For the elements within the webpage HTML, we designed several rules for filtering and retaining them. This process allowed us to preserve elements that are likely to be interactive components. At this stage, we save webpage screenshots, element information, and layout graphs (with different types such as interactive text buttons, interactive icon buttons, and images corresponding to specific colors). After this stage, there retained 285M webpages.

\paragraph{Rule-based filtering} Subsequently, we designed more fine-grained filters and deduplication techniques at the webpage and element levels based on the preserved webpages. These filters eliminated many erroneous and overly simplistic webpages. After this phase, 73.5M webpages remained.

\paragraph{Element selection and labeling} Finally, when selecting elements, we consider the distribution of element centroids and their types. Specifically, we discretize and uniformly sample across various regions of the canvas. During sampling in a discrete area, we prioritize sampling icon elements, as they are less frequent. We sample 10 samples per screenshot and use GPT-4o to label the captioning of each element. 

\paragraph{Prompting o3} Then we put the annotated document to o3 to label tasks with the following system prompt.

\begin{tcolorbox}[title={System prompt}]
\begin{lstlisting}
You are a training data construction expert for a computer-
using intelligent agent. 
You will be provided with: a screenshot, and several 
elements within the screen (such as buttons, etc.) 
along with the click coordinates of these elements (usually
the center position of the element). 
Your task is to construct several screen operations that can
be performed on the current screen and provide the 
corresponding responses. 

Here are the detailed requirements:
# Training data

A piece of training data consists of:
- "prompt": A request for an operation that can be executed
on the current screen.
- "response": The expected reply from the agent, which 
includes an analysis of the prompt and PyAutoGUI code. In 
the code, all coordinates are replaced with symbols like x1,
y1, x2, y2, etc.
- "coordinate_map": A mapping dictionary from the substitute 
symbols to the actual values, such as {"x1": 0.12345,...}. 
It can be empty {} if the action do not contain coordinate 
in the provided elements.
- "used_elements": a list of the index of elements (which 
will be provided in the input as "element_id") for this 
data, it can be empty [] if the action do not contain 
coordinate in the provided elements.
- "action-types": str, the action type name, e.g., "scroll", 
"moveTo", "click", "combined:mouseDown moveTo and mouseUp"
A complete piece of data is constructed with these four 
components forming a dictionary

# Task requirements

There are 3 classes of action types, classified by the 
number of coordinates used and the PyAutoGUI action type:
ZeroSet: "scroll", "typewrite", "hotkey"
OneSet: "moveTo", "click", "mouseDown", "mouseUp"
TwoSet and combined: for example, "combined:moveTo and 
dragTo", "combined:mouseDown moveTo and mouseUp", 
"combined:click and type"
We will introduce how to build them.

## ZeroSet task

It means the task do not need to use any coordinates in the 
Input, the used\_elements and coordinate_map should be 
empty.


\end{lstlisting}
\end{tcolorbox}

\begin{tcolorbox}[title={System prompt}]
\begin{lstlisting}
An example result for "scroll":

{
    "prompt": "I want to find xxx but I do not see xxx in 
    this screen, can you scroll to find them?",
    "response": "Let's see the screen ..., ... so I should 
    scroll down for the user 
    \n\n```python\n\npyautogui.scroll(-100)\n```",
    "coordinate_map": {},
    "used_elements": [],
    "action-type": "scroll"
}
when you generate the data, remember to cover the parameters
of the PyAutoGUI function with diverse setting (like you can 
say scroll -200 in prompt and do so in response).

## OneSet task

Take click as example, assume there is a 'update' button in 
input {"element_id": 3, "description": "a 'update' button",
"click_point": (0.12345, 0.67891)}

You may generate:
{
    "prompt": "Right click twice on the button that can help 
    me update the xxx",
    "response": "I can see ..., ... so I should right click
    on it twice: \n\n```python\n\npyautogui.click(x=x1, 
    y=x2, clicks=2, button='right')\n```",
    "coordinate_map": {"x1": 0.12345, "x2": 0.67891},
    "used_elements": [3],
    "action-type": "click"
}
Also remember to be diverse across all parameters (left, 
right click, num clicks, etc.)

## TwoSet and combined
Sometimes when we use computer several 2-3 actions are 
clustered with semantics, like when you type on text box, 
usually you need to click to focus first.
Some commonly used are: "combined:moveTo and dragTo", 
"combined:mouseDown moveTo and mouseUp" for selecting text 
or drag something,
"combined:click and type" for text box focus and type
"combined:moveTo and scroll" for sub-container scrolling, 
etc.

However, you job is to maximize the diversity of actions 
combination, instead of making the most appropriate data, 
which means even the button cannot be drag, you can drag, 
even the region can not type, you can also click and type.



\end{lstlisting}
\end{tcolorbox}

\begin{tcolorbox}[title={System prompt}]
\begin{lstlisting}
here is an example:
{
    "prompt": "Please drag the button of xxx to the position 
    of xxx button.",
    "response": "I can see ..., ... so I should drag from xx 
    to xx: \n\n```python\n\npyautogui.moveTo(x=x1, 
    y=y1)\npyautogui.dragTo(x=x2, y=y2)\n```",
    "coordinate_map": {"x1": 0.12345, "y1": 0.67891, "x2": 
    0.24732, "y2": 0.94724},
    "used_elements": [3, 7],
    "action-type": "combined:moveTo and dragTo"
}


# Styling
**IMPORTANT: the above example is only to help you 
understand the task. For the style and other requirements of
the prompt and response, please refer to the following 
standards!!! **

## prompt
- The style of the prompt should be diverse. It can be a 
direct and clear request, such as "type xxx". It can also be
a first-person question with user context, or it can be a 
vague request. For example, if there is a magnifying glass-
like icon on the screen, the prompt could simply say "help 
me enlarge this image" (rather than saying to click the 
magnifying glass icon button).
- If the corresponding function in PyAutoGUI has parameters,
the task needs to cover as many parameters as possible. 
However, the prompt and response must be consistent. For 
example, if the prompt specifically requests a right-click 
(left-click can be default), the response should also 
include a right-click.
- If you find a nice chance to generate combined actions, 
don't miss it!

## responses
The format of the response must be as follows:
"<chain of thought>\n\n```python\n<pyautogui code>\n```"
- In the python part, you don't need to import pyautogui, 
and you are NOT allow to generate any code that is not a 
pyautogui action. No comments allowed!
- The chain of thought part should be insightful for the 
prompt, and should be at least 2-3 sentences, more but 
useful is better.


# Number requirements

You need to generate **10** data in total, in which at least
**5** must be OneSet task, and the other can be set 
dynamically according to the situation.

\end{lstlisting}
\end{tcolorbox}

\begin{tcolorbox}[title={System prompt}]
\begin{lstlisting}
\# Overall formatting

Input format: A screenshot and a list of dictionaries with 
keys "element_id", "description", and "click_point".

Output format: A list of output dictionaries as described 
above with the following JSON format
```json
[
    {
        "prompt": ...,
        "response": ...,
        "coordinate_map": ...,
        "used_elements": [...]
    },
    ...
]
```
\end{lstlisting}
\end{tcolorbox}

\begin{figure}[H]
    \centering
    \includegraphics[width=\textwidth,trim=0 50 0 0,clip]{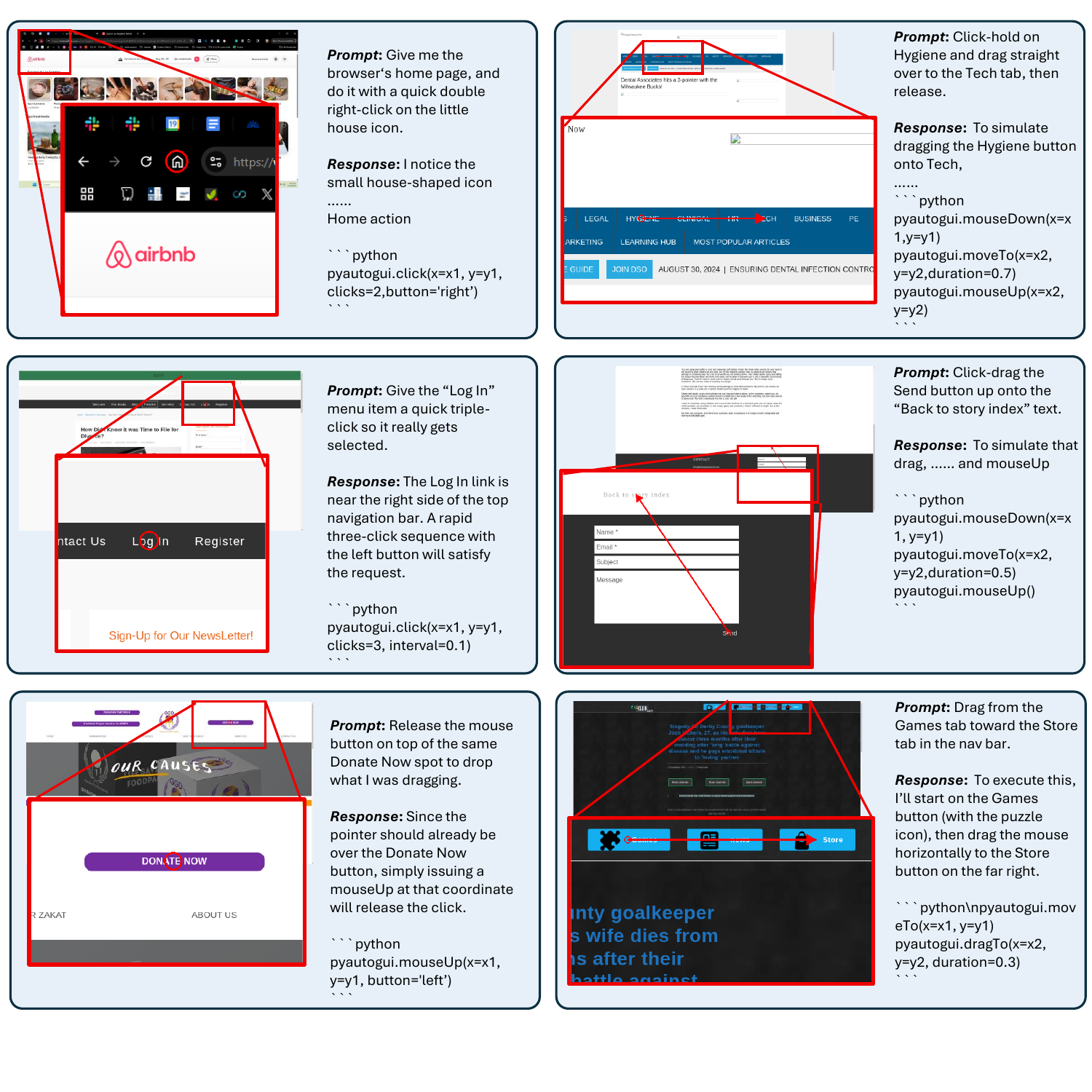}
    \vspace{-0.3cm}
    \caption{Examples of GUI modal data.}
    \label{fig:visual-gui}
\end{figure}

\subsection{Text}
The processing of the text modality differs slightly from that of the other modalities. If coordinates were recorded for every individual character and then generated by the model, the annotation files would become excessively large. In practice, for the text modality, we consider only six scenarios: two data types: code and natural language, and three task types: drag, selecting a short text span, drag-selecting a long text span, and clicking to place the insertion cursor.
We distinguish between short and long text selection because the corresponding references often differ. In particular, short text spans may introduce ambiguity, as the same text can appear multiple times within a document; consequently, selecting a short span may require additional contextual information. After manually identifying the target region, we prompted GPT to reformulate the task descriptions. Samples of the resulting data are shown in the figure below.

\begin{figure}[H]
    \centering
    \vspace{-0.3cm}
    \includegraphics[width=\textwidth,trim=0 45 0 0,clip]{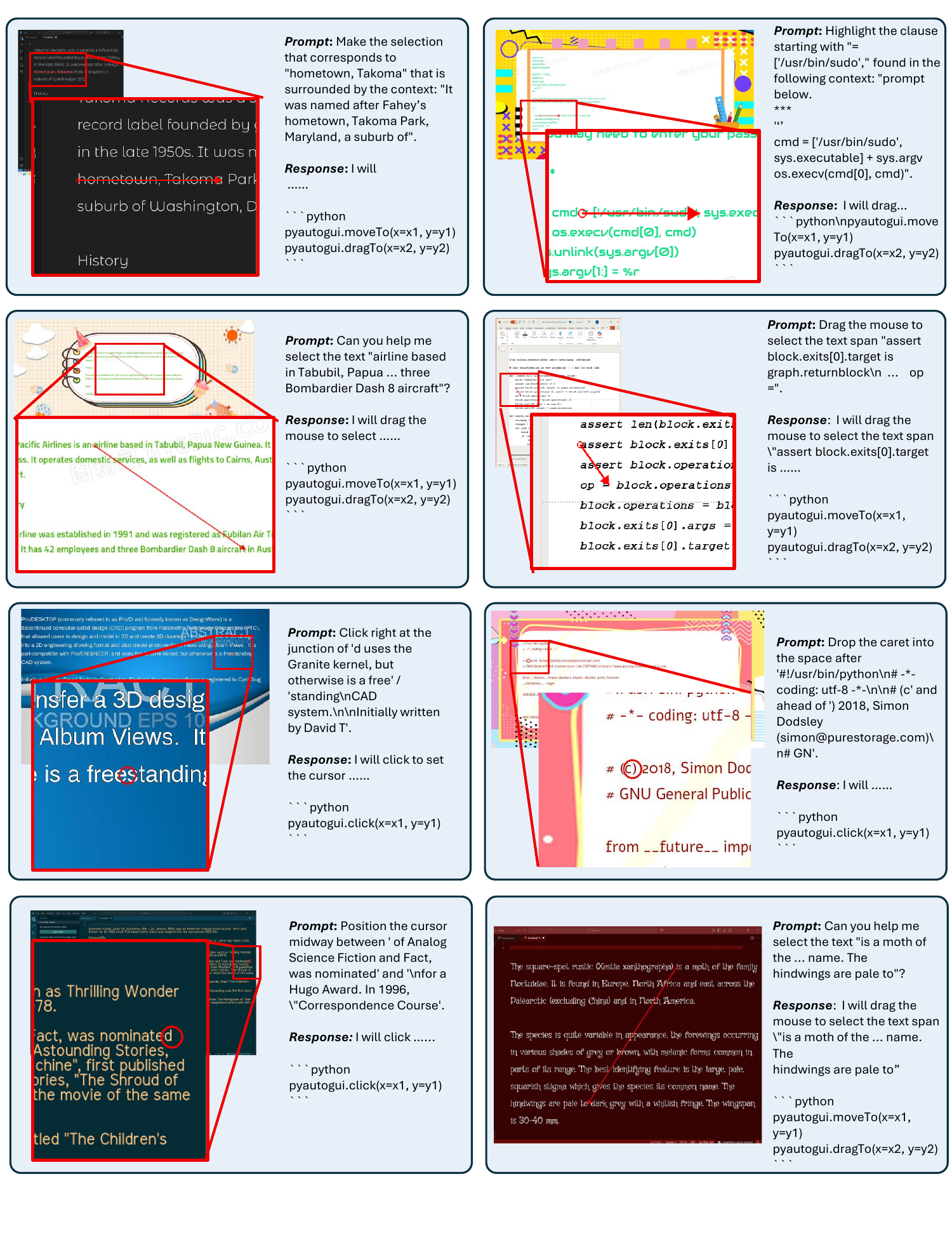}
    \vspace{-0.3cm}
    \caption{Examples of Text modal data.}
    \label{fig:visual-text}
\end{figure}

\subsection{Table}
The rendering of table data is divided into the following four stages:
\begin{itemize}
    \item \textbf{Step 1}: Collect open-source TableVQA-related datasets from Hugging Face and other sources. These datasets are often accompanied by table information in Markdown, HTML, or arXiv formats. This resulting about 16k tables.
    \item \textbf{Step 2}: use GPT to re-gen based on the seed table: (1)
changing topic (like change the table of working hour
to the statistic of math scores) and (2) Changing
topology (like adding a new merged cell and a new line)
This result in about 16k $\times$ 10 = 160k unique table data.
    \item \textbf{Step 3:} Using o3, we generated style sheets in diverse visual styles based on various open-source CSS libraries. The parameters of each style sheet—such as color, font size, cell size and type, and the presence or absence of borders—were designed to be adjustable and randomly sampled. In the end, we created 1k templates, and for each template, we sampled 10 different parameter configurations, resulting in a total of 10k style sheet instances.
    \item \textbf{Step 4:} By randomly combining HTML tables with CSS attributes, we can obtain complete table webpages. In practice, we further select half of the tables and randomly mask out most of their cells. This setting is common in source tables from applications such as Excel, where a large number of empty cells substantially increases the difficulty of both cell grounding and reference generation.
\end{itemize}
We have now successfully obtained the table HTML, and with JavaScript, we can readily render the bounding-box coordinates of each cell. To prevent the model from hallucinating the row index or positional information when generating tasks, we also use code to compute, for each element, its row number, column number, the corresponding row and column headers, and the cell content, which are then provided as references to the model. 
\begin{figure}[H]
    \centering
    \vspace{-0.2cm}
    \includegraphics[width=\textwidth,trim=0 170 0 0,clip]{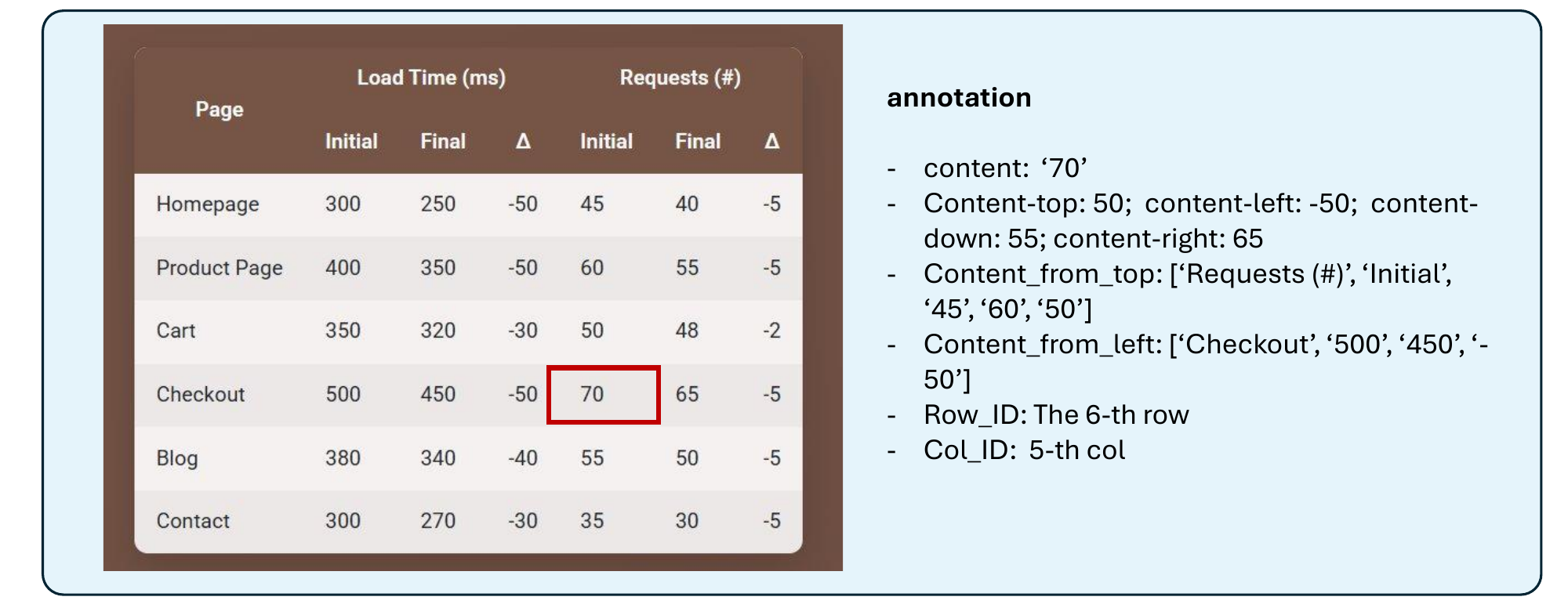}
    \vspace{-0.5cm}
    \caption{Examples of annotation of table cell.}
    \label{fig:table-ann}
    \vspace{-0.5cm}
\end{figure}

Figure \ref{fig:table-ann} illustrates the complete annotation of a single cell. We feed a screenshot, along with multiple sampled cells and their annotations, into o3, and use the following system prompt to generate the final tasks.

\begin{tcolorbox}[title={System prompt}]
\begin{lstlisting}
You are a training data construction expert for a computer-
using intelligent agent. You will be provided with: a 
screenshot, and a cell and its information of a table within 
the screen.

\end{lstlisting}
\end{tcolorbox}

\begin{tcolorbox}[title={System prompt}]
\begin{lstlisting}
Your task is to construct several GUI operations that can be 
performed on the current screen and provide the corresponding 
responses. 

# Training data

......
[Same with GUI]
......

## OneSet task

......
[Same with GUI]
......


Hints: Also remenber that the coordinate in the training data
may be computed from the given information. 
Besides using cell level and compute center point, you may 
also generate tasks like:
- drag the right edge of xxx cell to 10 px right to wider the 
whole column (usually seen in apps like Excel) and you compute
the middle of right edge (x1, y1) = (X2, (Y1+Y2)/2) and (x2, 
y2) = (x1+10, y1).
- if in the given screen, cell A (X1, Y1, X2, Y2) is right 
above the cell B (when the size of cells are similar, you can 
compute the coordinate of B is (X1, Y2, X2, 2*Y2-Y1)), you can 
generate task like drag the right-bottom corner of cell A to 
the right-bottom corner of cell B. In Excel, this usually 
means transfer the formula to cells between A and B, you can 
do so even though the provided image may not be Excel.
- ......
Please use your imagination to generate various rich data.

# Styling
**IMPORTANT: the above example is only to help you understand 
the task. For the style and other requirements of the prompt 
and response, please refer to the following standards!!! **

## prompt
- The style of the prompt should be diverse. It can be a 
direct and clear request, such as "type xxx". It can also be a 
first-person question with user 
- If the corresponding function in PyAutoGUI has parameters, 
the task needs to cover as many parameters as possible. 
However, the prompt and response must be consistent. For 
example, if the prompt specifically requests a right-click 
(left-click can be default), the response should also include 
a right-click.
\end{lstlisting}
\end{tcolorbox}

\begin{tcolorbox}[title={System prompt}]
\begin{lstlisting}
- If you find a nice chance to generate combined actions, 
don't miss it!
- Ensure that the prompt do NOT have ambiguity, for example, 
there may be severl empty cell or cell with content '3', if 
this happens, refer clearly which cell (you can use 'the empty 
cell in A-3', 'the number 3 under the header apple', etc)
- Double check and make sure the computation of coordinates 
are correct!


## responses
The format of the response must be as follows:
"<chain of thought>\n\n```python\n<pyautogui code>\n```"
- In the python part, you don't need to import pyautogui, and 
you are NOT allow to generate any code that is not a pyautogui
action. No comments allowed!
- The chain of thought part should be insightful for the 
prompt, and should be at least 2-3 sentences, more but useful 
is better.
- **NO any coordinate can be appear in the chain of thought 
part!!!** Because in the future, we may do data argmentation 
on the coordinates, that why we use coordinate map. 


# Number requirements

You need to generate **4** data in total, in which at least 
**1** must be OneSet task, and the other can be set 
dynamically according to the situation.

# Overall formatting

Input format: A screenshot and a dictionary with keys 
"description", and "bbox".

Output format: A list of output dictionaries as described 
above with the following JSON format
```json
[
    {
        "prompt": ...,
        "response": ...,
        "coordinate_map": ...,
    },
    ...
]
```

\end{lstlisting}
\end{tcolorbox}

The final tasks generated by the model encompass not only clicking operations, but also actions such as dragging cells and adjusting cell boundaries. The figure below presents several examples of real table-manipulation tasks produced by our pipeline.

\begin{figure}[H]
    \centering
    \vspace{-0.2cm}
    \includegraphics[width=\textwidth,trim=0 0 0 0,clip]{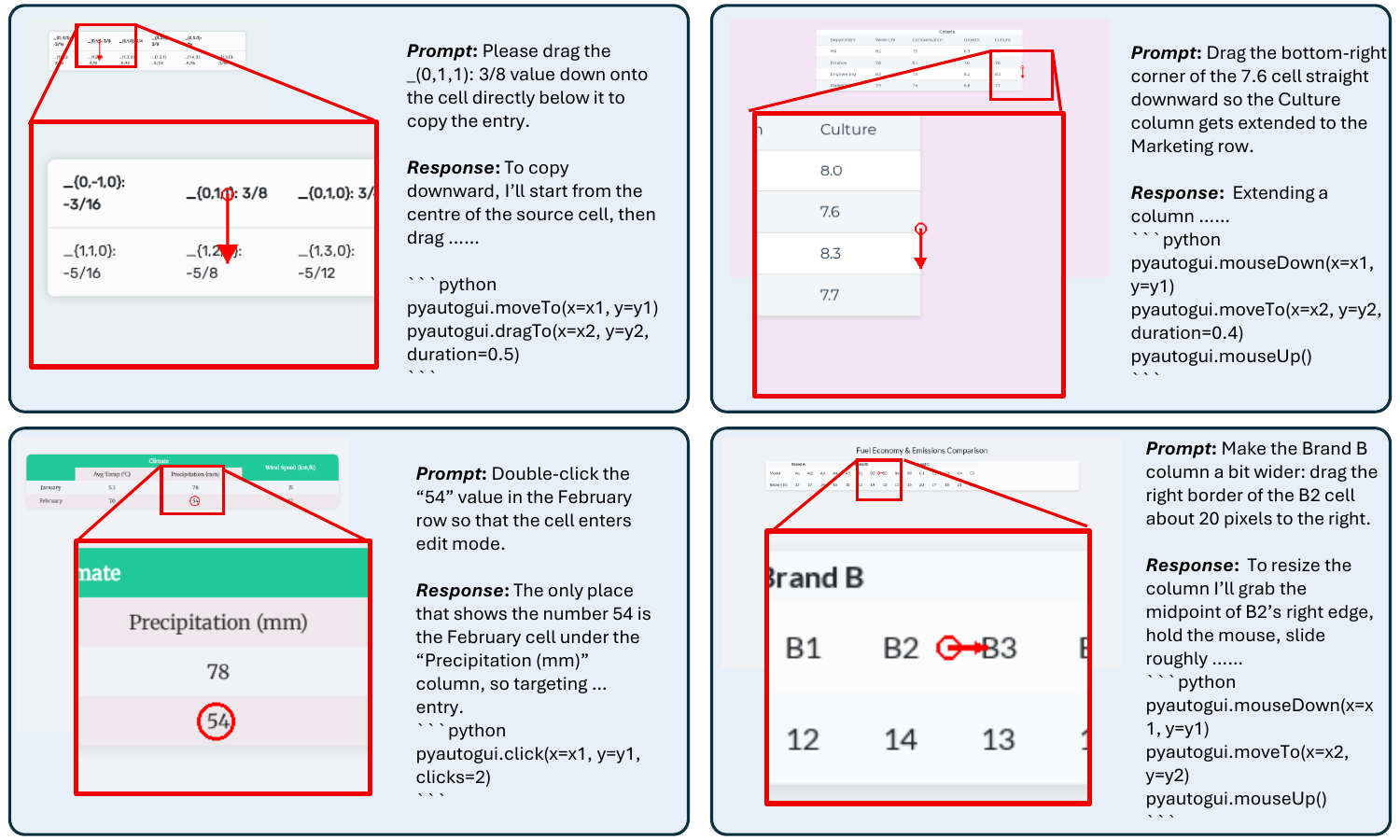}
    \vspace{-0.5cm}
    \caption{Examples of Table data.}
    \label{fig:table-visual}
    \vspace{-0.5cm}
\end{figure}

\subsection{Canvas}

To bootstrap a large-scale corpus for visual action grounding, we
implement a fully procedural \emph{PowerPoint-style canvas simulator}
that renders raster scenes resembling slides under active editing,
together with rich geometric annotations.  Every image is paired with a
structured JSON label that exposes the bounding box, center, vertices,
endpoints, eight bounding-box control points, and the rotation handle of
each rendered element, so that downstream PyAutoGUI-style operation
trajectories can be constructed without any human annotation.

\paragraph{Canvas and Element Sampling}
Each scene is parameterized by a randomly sampled canvas size
$W\!\in\![800, 2560]$, $H\!\in\![600, 1440]$, and a background color
drawn in HSV space.  Between $3$ and $8$ elements are then placed on the
canvas.  Element types are sampled with importance weighting that
slightly favors \emph{common} primitives (rectangle, circle, triangle,
star, diamond, basic arrows) over the rarer ones to mimic typical slide
distributions.  Sizes range from $8\%$ to $40\%$ of the shorter canvas
side; line-like elements are allowed to span up to $60\%$.

\paragraph{Overlap-aware placement.}
Elements are placed sequentially with up to $50$ random trials per item.
A candidate bounding box is accepted when its maximum pairwise overlap
ratio with all previously placed boxes (relative to the smaller area) is
below $0.25$; otherwise the lowest-overlap candidate is retained as a
fallback.  Square-aspect shapes (circle, square, donut, ring,
rounded square) are constrained to equal width and height to preserve
geometric semantics.

\paragraph{Color model.}
Background, fill, and outline colors are drawn from HSV with rejection
sampling against a redmean-weighted Euclidean distance metric, enforcing
minimum perceptual gaps of $100$ from the background and $60$ between
fill and outline.  Outlines are additionally randomized between solid
and dashed strokes (with $0.2$ probability of dashing) and between $1$
and $5$~px widths.

\paragraph{Shape Library}
The simulator exposes a registry-based shape library covering
\textbf{$76$ primitive types} grouped into nine categories:
\begin{itemize}
\item \textbf{Rectangles}: rectangle, rounded rectangle, square,
  rounded square, cross, plus.
\item \textbf{Ellipses}: ellipse, circle.
\item \textbf{Triangles}: scalene, right, isosceles, equilateral, and
  obtuse triangles.
\item \textbf{Quadrilaterals}: diamond, parallelogram, trapezoid,
  right trapezoid, kite.
\item \textbf{Polygons}: pentagon, hexagon, heptagon, octagon, nonagon,
  decagon (procedurally generated for any $n$).
\item \textbf{Stars}: $4$-, $5$-, $6$-, $8$-, $10$-, $12$-point stars.
\item \textbf{Arrows}: right/left/up/down/double-headed arrows,
  chevrons, notched arrow, bent arrow, U-turn arrow, circular arrow.
\item \textbf{Lines and connectors}: straight line, single-arrow line,
  double-arrow line, curved (B\'ezier) line, elbow connector.
\item \textbf{Callouts and decorations}: rectangular callout, rounded
  callout, cloud callout, ribbon, banner.
\item \textbf{Special / decorative shapes}: heart, cloud, crescent moon,
  sun, frame, donut, ring, lightning bolt, wave, arc, pie, sector,
  drop, explosion, semicircle, quarter circle, teardrop, shield,
  L-shape, T-shape.
\item \textbf{Text boxes}: bordered, rounded-border, and borderless
  text boxes.
\end{itemize}
All shapes implement a common \texttt{ShapeDrawer} interface that
returns a \texttt{ShapeResult} dataclass containing the bounding box,
center, named vertices, and (for line-like shapes) endpoints, allowing
uniform downstream serialization.  Polygons support both solid and
dashed strokes via a custom dashed-polygon rasterizer; arrows are drawn
with parametric arrowheads of configurable size and direction.

\paragraph{PPT-style Selection Markers}
To mimic an authoring environment, each element is overlaid with the
selection chrome of a typical slide editor:
(i) a thin gray bounding box;
(ii) eight red \emph{control points} at the four corners and four edge
midpoints (or at the two endpoints for line-like elements);
(iii) blue diamond \emph{vertex markers} at every named polygon vertex
(skipped for shapes whose ``vertices'' are dense curve approximations,
e.g.\ heart, cloud, moon, wave);
(iv) a \emph{rotation handle} consisting of a short connector and a
$300^\circ$ circular arrow with a small arrowhead tip, randomly anchored
to one of the four bounding-box midpoints.
The exact pixel coordinates of every marker are recorded in the
annotation, providing fine-grained, action-ready interaction targets
beyond the geometric center.

\paragraph{Reference Expression Generation}
Each element is paired with a unique English referring expression of
the form
\emph{``\textless fill-color\textgreater-filled \textless shape\textgreater\
with \textless outline-color\textgreater\ outline in the
\textless region\textgreater''}.
Color words are obtained by nearest-neighbor lookup against a
$44$-entry named-color palette using the redmean distance.  The canvas
is partitioned into a $3\!\times\!3$ region grid (e.g.\ ``upper-left
area of the canvas'').  When two elements collide in the base
description, a cascade of disambiguation strategies is applied in
order: relative-size descriptors (\emph{the largest / smallest / a
larger / a smaller}), line-style qualifiers (\emph{solid / dashed
outline}), refinement to a finer $5\!\times\!5$ region grid, and
finally a reading-order ordinal prefix (\emph{the upper, the second,
\ldots, the lower}).  This guarantees that every element in a scene
admits at least one unique natural-language reference.

\paragraph{Annotation}
For every generated image the simulator emits a JSON file whose
\texttt{elements} array contains, per shape:
the unique id, the symbolic \texttt{shape\_type}, the disambiguated
\texttt{reference} string, the \texttt{bbox}, the \texttt{center\_point},
the eight named \texttt{box\_points} (\emph{top\_left}, \emph{top\_center},
\ldots, \emph{left\_center}), the \texttt{rotation\_handle\_center}, the
full styling dictionary (fill, outline, stroke width, line style), and
optional \texttt{vertices} or \texttt{endpoints} dictionaries for
polygonal and line-like shapes.  This schema is consumed downstream as
the structured input for the o3 model, with the following system prompt.

\begin{tcolorbox}[title={System prompt}]
\begin{lstlisting}
You are a training data construction expert for a computer-
using intelligent agent. 
You will be provided with: a screenshot, and several 
elements within the screen (such as shapes, arrows, etc.) 
along with the control point coordinates of these elements
(e.g., center point, top-left scaling point, arrow point, 
etc. Usually seem in PPT). 
\end{lstlisting}
\end{tcolorbox}

\begin{tcolorbox}[title={System prompt}]
\begin{lstlisting}
Your task is to construct several screen operations that 
can be performed on the current screen and provide the 
corresponding responses. 
Here are the detailed requirements:

# Training data
......
[Same with GUI]
......
## OneSet task
Take "combined:click and type" as example, assume there is
a square shape in input {"id": "shape_0001", "shape_type":
"rounded_square", "reference": "dark gray-filled rounded 
square with tan outline in the center-right area of the 
canvas", "center_point": [672, 608], "box_points": 
{"top_left": [548, 484],......}}
You may generate:
{
    "prompt": "left click twice on the dark gray-filled 
    rounded square to focus and then type 'ABC' in it",
    "response": "I can see ..., ... so I should left click 
    on it twice and then type:```python\n\npyautogui.
    click(x=x1, y=y1, clicks=2)\npyautogui.type('ABC')\n```",
    "coordinate_map": {"x1": 672, "y1": 608},
    "used_elements": ["shape_0001"],
    "action-type": "combined:click and type"
}
Also remember to be diverse across all parameters (left, 
right click, num clicks, etc.)
## TwoSet and combined
Sometimes when we use computer several 2-3 actions are 
clustered with semantics.
Some commonly used are: "combined:moveTo and dragTo", 
"combined:mouseDown moveTo and mouseUp" for dragging 
something.

You job is to maximize the diversity of actions 
combination, and control point use.
# Styling
**IMPORTANT: the above example is only to help you 
understand the task. For the style and other requirements 
of the prompt and response, please refer to the following 
standards!!! **
## prompt
- The style of the prompt should be diverse. It can be a 
direct and clear request, such as "click and type xxx". It 
can also be a first-person question with user context 
(e.g., 'Now you have selected an arrow drawing tool, please 
drag from the left control point to ...'), or it can be a 
vague request. 
- If the corresponding function in PyAutoGUI has 
parameters, the task needs to cover as many parameters as 
possible. However, the prompt and response must be 
consistent. For example, if the prompt specifically 
requests a right-click (left-click can be default), the 
response should also include a right-click.

\end{lstlisting}
\end{tcolorbox}

\begin{tcolorbox}[title={System prompt}]
\begin{lstlisting}

- If you find a nice chance to generate combined actions,
don't miss it!
- You are encouraged to do calculation based on the 
provided info. For example, you can generate task like 
'drag the bottom-right scaling control point of xxx to make 
it a 30px larger in both w and h.' and use coordinate (x, 
y) and (x+30, y+30).
Or you can let terms like 'the middle of the right side of
shape A and the left side of shape B', and then you 
calculate the avg of two coordinates.
## responses
The format of the response must be as follows:
"<chain of thought>\n\n```python\n<pyautogui code>\n```"
- In the python part, you don't need to import pyautogui, 
and you are NOT allow to generate any code that is not a 
pyautogui action. No comments allowed!
- The chain of thought part should be insightful for the 
prompt, and should be at least 2-3 sentences, more but 
useful is better. BUT do NOT include any coordinates in the
chain of thought part.


# Number requirements
You need to generate **10** different data in total. At 
most 3 use the center point, at most 4 use other control 
points (others use calculated coordinate)

# Overall formatting
......
[Same with GUI]
......
\end{lstlisting}
\end{tcolorbox}

\begin{figure}[H]
    \centering
    \vspace{-0.2cm}
    \includegraphics[width=\textwidth,trim=0 0 0 0,clip]{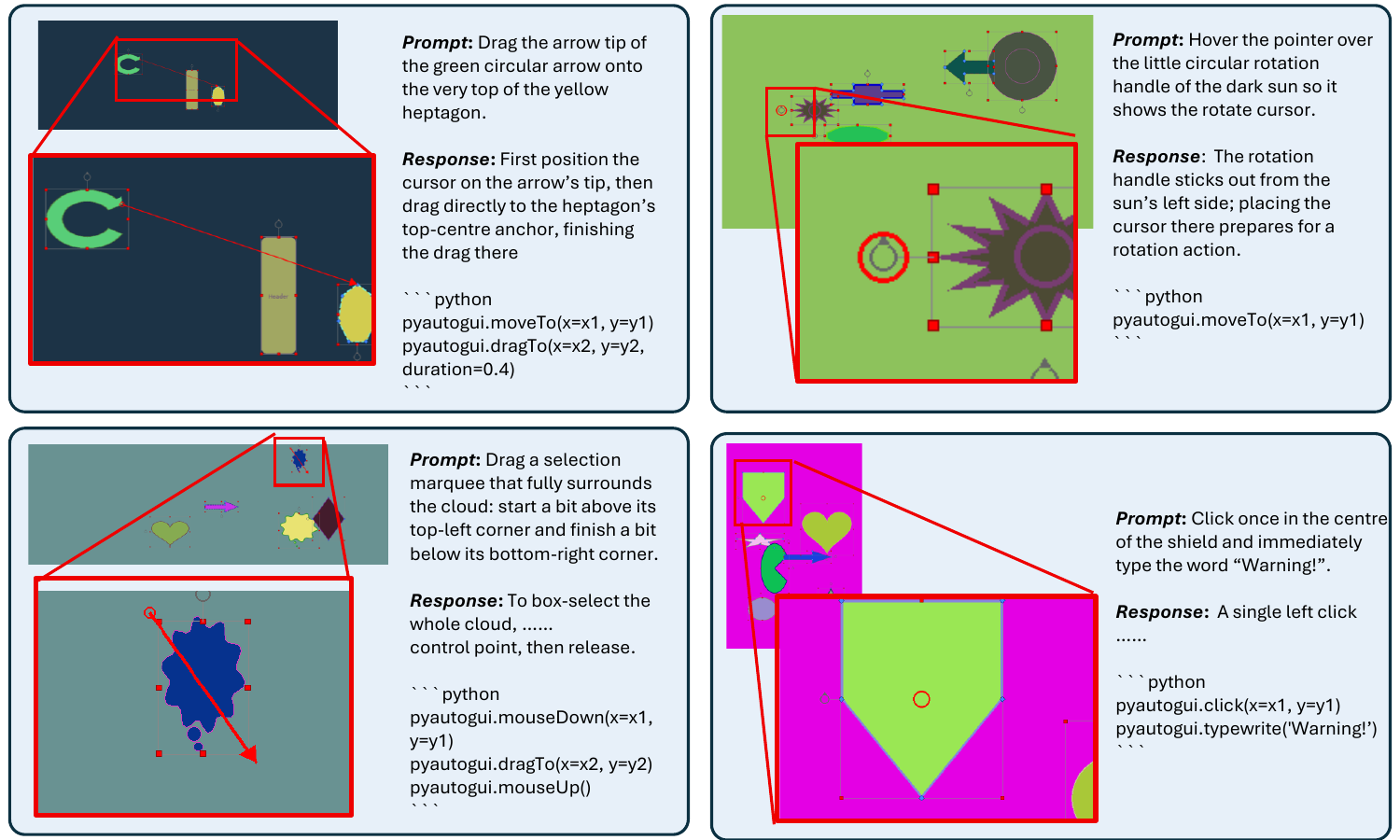}
    \vspace{-0.5cm}
    \caption{Examples of Canvas data.}
    \label{fig:table-visual}
    \vspace{-0.5cm}
\end{figure}

\subsection{Natural Image}
For natural images, we directly sample instances from the SAM dataset. For each image, we select five regions and use GPT-4o to generate captions for them. As a result, each region is associated with several attributes for reference by o3, including a caption, bounding box, boundary, and center point. Specifically, the boundary is represented as a 20-point polygon derived from the region mask, while the center point is computed from the bounding box. We then provide o3 with the image together with the annotations of the five regions, and prompt the model using the following system prompt.

\begin{tcolorbox}[title={System prompt}]
\begin{lstlisting}
You are a training data construction expert for a computer-
using intelligent agent. 
You will be provided with: a natural image, and several 
region within the screen
along with many information about the region: bbox, center 
point, boundary polygon, etc. 
The bbox and boundary of each element will also be mark on 
the image for you to recognize.

Your task is to construct several mouse operations that can
be performed on the current image (this happens offently when
you usse apps like PS, GIMP meitixiuxiu etc.) and provide the
corresponding responses. 
Here are the detailed requirements:

# Training data
......
[Same with GUI]
......
## OneSet task
Take click as example, assume there is a 'Geep car' in image
{"id": 3, "caption": "the blue polygon shows a 'Geep car'", 
"center point": (123, 456)}
You may generate:
{
    "prompt": "You are using a auto-selection crop function, 
    please right click on the car to select it.",
    "response": "I can see ..., ... so I should right click 
    on it: \n\n```python\n\npyautogui.click(x=x1, y=x2, 
    clicks=1, button='right')\n```",
    "coordinate_map": {"x1": 123, "x2": 456},
    "used_elements": [3],
    "action-type": "click"
}
Remember to be diverse across all parameters (left, right 
click, num clicks, etc.)
You can use every information you can find in the information 
dict, not just center point. 
For example, you can use the boundary polygon, if you see the
left is the car's head and you can use the most left point in
the given boundary, to generate a task like click on the head
of the task.

\end{lstlisting}
\end{tcolorbox}

\begin{tcolorbox}[title={System prompt}]
\begin{lstlisting}
## TwoSet and combined
This means we need at least 2 coordinates for a task.
This can be use the bbox of one region: task like 'drag a 
bounding box for the red house on the left of the image'
You can also use two coordinates of the boundary of one 
element like 'drag an arrow from the head of the car to the 
tail' You can also use coordinates from two regions.
You can also do calculation, like assuming you are using a 
slim waist/face function, you will need to pick a point near 
the face and drag from outside of the boundary to inside.
here is an example:
{
    "prompt": "I have already crop the house in the image, 
    please drag the center of the house to the position of 
    the car to cover the car.",
    "response": "I can see ..., ... so I should drag from xx 
    to xx: \n\n```python\n\npyautogui.moveTo(x=x1, 
    y=y1)\npyautogui.dragTo(x=x2, y=y2)\n```",
    "coordinate_map": {"x1": 345, "y1": 891, "x2": 732, "y2":
    724},
    "used_elements": [3, 5],
    "action-type": "combined:moveTo and dragTo"
}

## NSet
This only contain 2 types of task: (1) drag the mouse to 
select the boundary of the region. (2) draw to fill the 
content of the region.
This 2 task is usually used for cropping or masking, eraing 
parts of the image.
The boundary polygon in the information is an ordered list of 
points. If the region' space is large, you can just use that,
if the space is small, you can sample some of the points in 
order.
For the second task, you can change the order of the 
boundary, then it will be a left right left right trail to 
mask the whole image:
e.g., the boundary is : [p1, p2, p3, ... , p19, p20], then 
use [p1, p2, p20, p3, p19, p4, p18, ...] will draw and mask
the reigion.
here is an example:
{
    "prompt": "Now I am using the image erase tool, I need to 
    draw to cover the region of the man in yellow.",
    "response": "I can see ..., ... so I should drag to ...: 
    \n\n```python\n\npyautogui.mouseDown(x=x1, 
    y=y1)\npyautogui.moveTo(x=x2, 
    y=y2)\n......pyautogui.moveTo(x=x13, 
    y=y13)\npyautogui.mouseUp()\n```",
    "coordinate_map": {"x1": 345, "y1": 891, "x2": 732, "y2":
    724......, "x13": 14, "y13": 532},
    "used_elements": [4],
    "action-type": "combined:mouseDown NxmoveTo and mouseUp"
}
\end{lstlisting}
\end{tcolorbox}

\begin{tcolorbox}[title={System prompt}]
\begin{lstlisting}
# Styling
**IMPORTANT: the above example is only to help you understand 
the task. For the style and other requirements of the prompt 
and response, please refer to the following standards!!! **

## prompt
- The style of the prompt should be diverse. It can be a 
direct and clear request. It can also be a first-person 
question with user context, or it can be a vague request. 
- If the corresponding function in PyAutoGUI has parameters, 
the task needs to cover as many parameters as possible. 
However, the prompt and response must be consistent. For 
example, if the prompt specifically requests a right-click 
(left-click can be default), the response should also include 
a right-click.
- Please put very clear context (like in the example, we say 
'Now I am using the image erase tool, ...' to explain why we 
need this action) and it should be diverse!
- There may be phrases like 'the red polygon / bbox' in the 
caption, but all keywords like 'the polygon' 'the blue bbox' 
is NOT allowed in the prompt, because that is draw for you to 
understand, but those polygon are NOT in the original natural 
image. 

## responses
......
[Same with GUI]
......
\end{lstlisting}
\end{tcolorbox}

\begin{figure}[H]
    \centering
    \includegraphics[width=\textwidth,trim=0 0 0 0,clip]{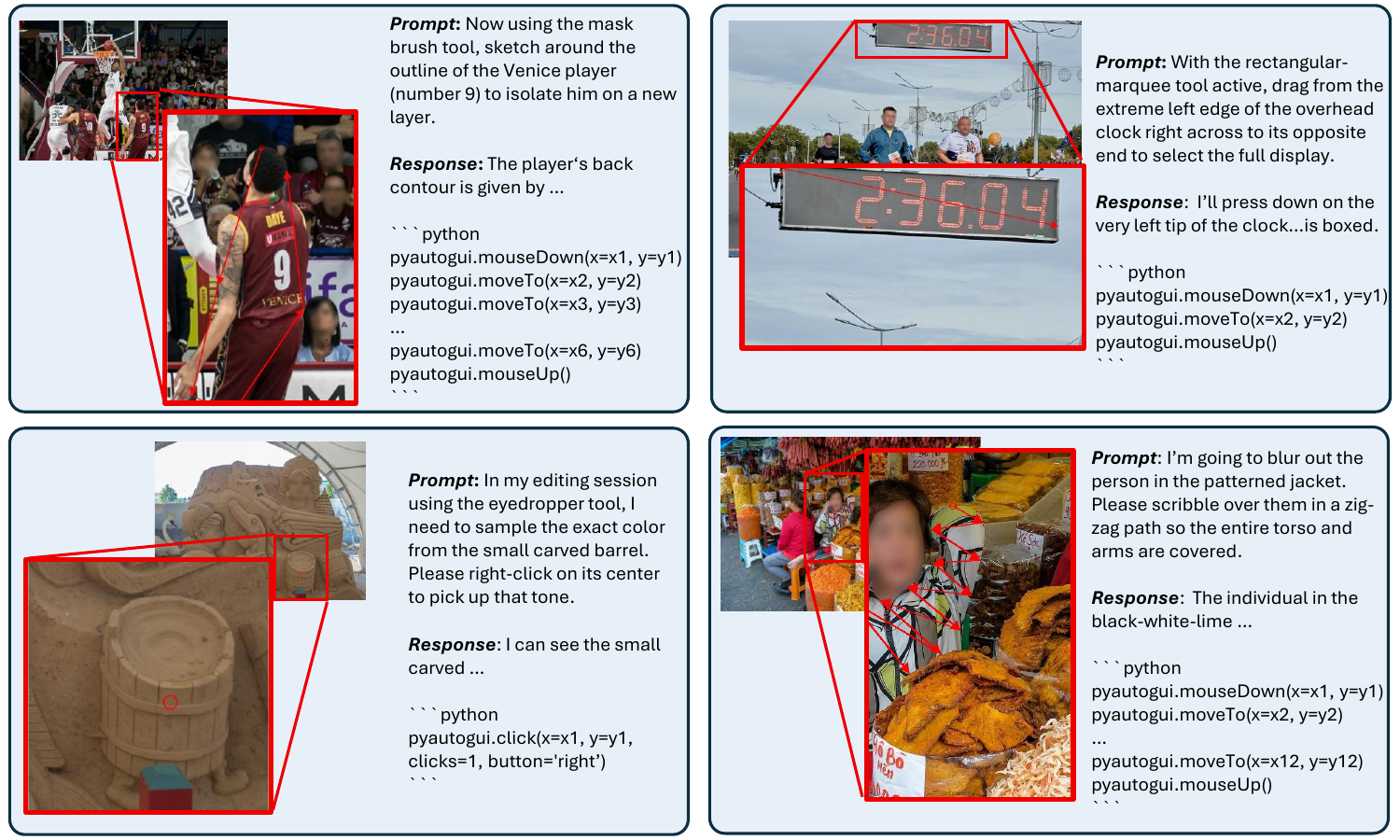}
    \caption{Examples of natural image grounding data.}
    \label{fig:image-visual}
\end{figure}

\section{Case study}
In the figure below, we present a drag-and-drop task from OSWorld. We use GPT-5.4 as the planner and Phi-Ground-Any as the grounder to execute the actions. As shown, from step 13 to step 29, the agent is required to repeatedly drag a specific cell to an empty cell below in order to copy its content into the blank cell. This relatively complex operation is executed correctly across multiple repetitions, demonstrating the drag-and-drop capability of Phi-Ground-Any on tabular data.

\begin{figure}[H]
    \centering
    \includegraphics[width=\textwidth,trim=0 100 0 0,clip]{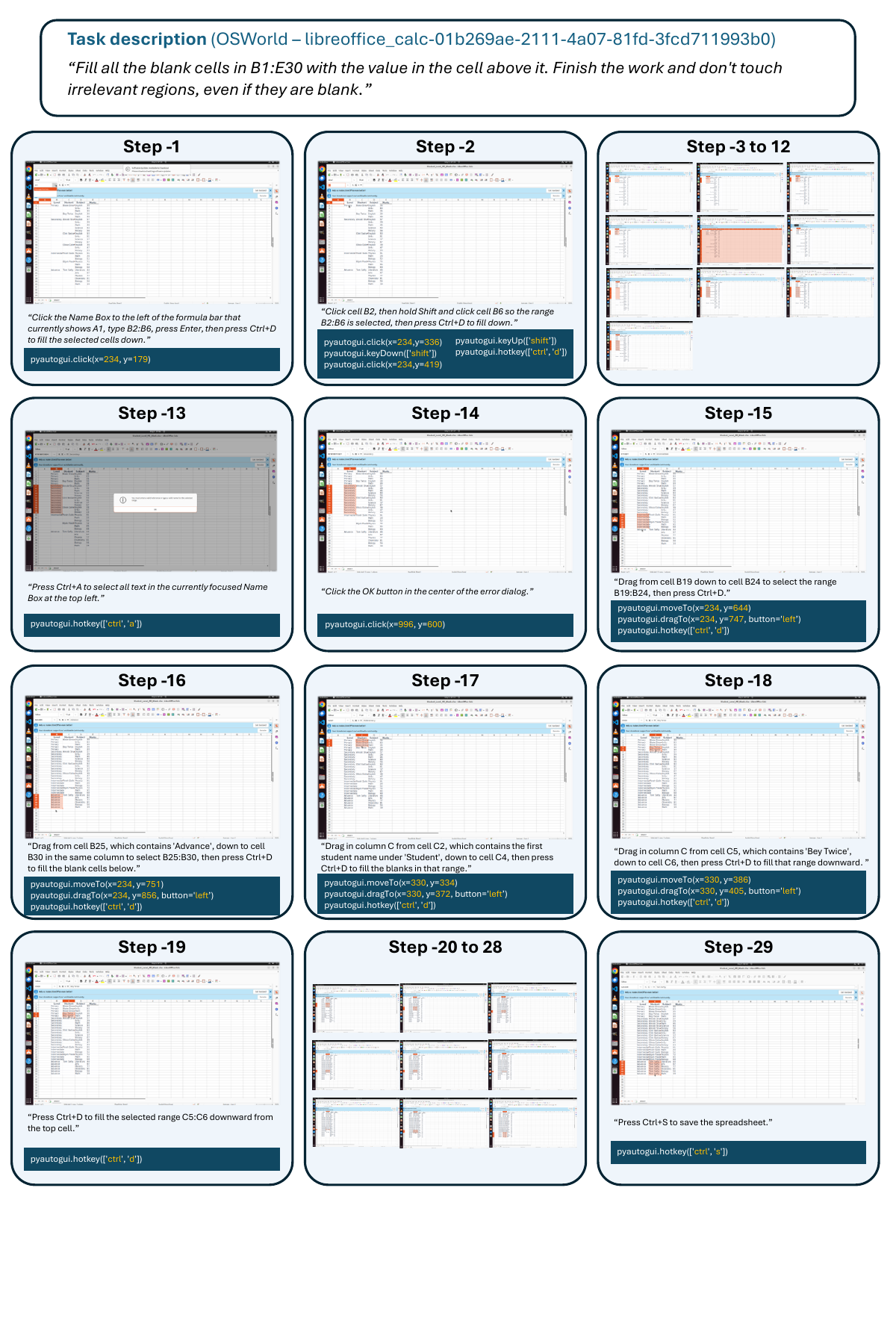}
    \vspace{-0.5cm}
    \caption{An case study of OSWorld Libreoffice-calc example.}
    \label{fig:case-study}
\end{figure}



\end{document}